\documentclass{article} % For LaTeX2e
\usepackage{iclr2026_conference,times}

\usepackage{amsmath,amsfonts,bm}

\def\eqref#1{equation~\ref{#1}}
\def\1{\bm{1}}

\DeclareMathAlphabet{\mathsfit}{\encodingdefault}{\sfdefault}{m}{sl}
\SetMathAlphabet{\mathsfit}{bold}{\encodingdefault}{\sfdefault}{bx}{n}

\usepackage{hyperref}
\usepackage{url}

\usepackage{soul}
\usepackage{url}
\usepackage[utf8]{inputenc}
\usepackage{graphicx}
\usepackage{amsmath}
\usepackage{amsthm}
\usepackage{booktabs}
\usepackage{algorithm}
\usepackage{algorithmic}
\usepackage[switch]{lineno}
\usepackage{amssymb}
\usepackage{pifont}
\usepackage{xcolor}
\usepackage{amssymb}
\usepackage{colortbl}
\usepackage[normalem]{ulem}
\useunder{\uline}{\ul}{}
\usepackage[inline]{enumitem}
\usepackage{enumitem}
\definecolor{customgreen}{rgb}{0.0, 0.5, 0.0}
\usepackage{arydshln}
\usepackage{tikz}
\definecolor{table_yellow}{RGB}{254,243,222}
\definecolor{table_blue}{RGB}{237,245,255}
\definecolor{table_green}{RGB}{234,245,234}
\usepackage{multirow}
\usepackage{wrapfig}
\usepackage{booktabs}
\usepackage{subcaption}
\usepackage{booktabs, graphicx, caption}

\title{Dynamic Multimodal Activation Steering for Hallucination Mitigation in Large Vision-Language Models}

\author{Jianghao Yin, Qin Chen\footnotemark[1] , Kedi Chen, Jie Zhou\footnotemark[1] , Xingjiao Wu, Liang He \\
East China Normal University \\
\texttt{jhyin@stu.ecnu.edu.cn, qchen@cs.ecnu.edu.cn, jzhou@cs.ecnu.edu.cn}\\
}

\iclrfinalcopy % Uncomment for camera-ready version, but NOT for submission.
\begin{document}

\maketitle
\renewcommand{\thefootnote}{\fnsymbol{footnote}}
\footnotetext[1]{Corresponding authors.}
\renewcommand*{\thefootnote}

\begin{abstract}
Large Vision-Language Models (LVLMs) exhibit outstanding performance on vision-language tasks but struggle with hallucination problems.
Through in-depth analysis of LVLM activation patterns, we reveal two key findings: 1) truthfulness and visual perception capabilities predominantly engage different subsets of attention heads within the model architecture; and 2) truthfulness steering vectors vary significantly across different semantic contexts. Based on these observations, we propose Dynamic Multimodal Activation Steering, a training-free approach for hallucination mitigation. Our method constructs a semantic-based truthfulness steering vector database and computes visual perception steering vectors, enabling context-aware interventions during inference by dynamically selecting the most relevant steering vectors based on input semantic similarity and applying them to the most influential attention heads. We conduct comprehensive experiments across multiple models and datasets, demonstrating that our approach significantly enhances model performance, outperforming existing state-of-the-art methods. 
\end{abstract}

\section{Introduction}
Large Vision-Language Models (LVLMs) have demonstrated remarkable performance on visual question answering (VQA), image captioning, and related tasks \citep{llava,llava1.5,qwenvl,internVL,instructblip}. However, these models suffer from significant hallucination phenomena \citep{huang2025survey, hallucination2}, manifested as fabricating non-existent objects or incorrectly describing image content \citep{lvlmhallusurvey, mllmhallusurvey}. Such hallucinations limit the applicability of LVLMs in downstream applications including autonomous driving \citep{autonomousdriving}, robotics \citep{robotic}, and other safety-critical domains.

Due to the complex architecture of LVLMs, the causes of multimodal hallucinations are diverse. To address these multimodal hallucination issues, numerous approaches have been proposed \citep{VCD,opera,AGLA,woodpecker,LRV}, which can be broadly categorized into two classes: training-based and decoding-based methods. Training-based methods primarily focus on constructing less biased datasets to fine-tune LVLMs, such as LRV \citep{LRV}, or employing reinforcement learning to train LVLMs, as demonstrated by RLHF-V \citep{rlhf}. The limitations of these approaches lie in their requirements for carefully curated data and substantial computational resources, as well as the need to retrain models separately for different architectures. Decoding-based methods, on the other hand, modify the decoding strategies of LVLMs, such as VCD \citep{VCD} and ICD \citep{ICD}. While these methods avoid the need for training, they often compromise the quality of the generated content \citep{VAF}.

More recently, researchers have begun investigating activation engineering \citep{representation, ITI, SADI} as an alternative approach to reduce hallucinations through targeted intervention in model representations. ICT \citep{ICT} is an image-object cross-level trusted intervention method that mitigates model hallucinations by applying noise to both images and objects, thereby enhancing the model's attention to visual information. However, this approach primarily focuses on visual-level interventions, neglecting the multimodal characteristics of LVLMs.
VTI \citep{VTI} intervenes in the hidden states of both the visual encoder and large language model during inference by pre-computing steering vectors for visual and textual modalities. Nevertheless, this method employs fixed steering vectors regardless of input variation, ignoring potential semantic differences across diverse contexts. The uniformly applied steering vectors fail to account for the nuanced semantic variations that exist across different inputs.

To address these challenges, we propose dynamic multimodal activation steering (DMAS), a training-free approach for hallucination mitigation in LVLMs. Our method focuses on two types of attention heads in LVLMs: truthfulness-related and visual perception-related. For truthfulness heads, we explicitly model how truthfulness steering vectors vary across semantic contexts. We cluster data semantically and create sample pairs with and without hallucinations within each cluster. By contrasting attention activations between factual and hallucination-prone samples, we extract truthfulness steering vectors. These vectors are stored alongside their cluster embeddings in a key-value database. For visual perception, we calculate activation differences between noise-free and noisy image inputs to derive perception steering vectors that enhance visual attention.
During inference, we dynamically retrieve the most semantically relevant truthfulness steering vector for the input query and apply both truthfulness and visual perception vectors to the top-K attention heads with the largest activation differences. This dual intervention effectively reduces hallucinations. The main contributions of our paper are:
\begin{itemize}[leftmargin=*, align=left]

\item We investigate activation differences in LVLMs, revealing that truthfulness and visual perception capabilities predominantly engage different subsets of attention heads, and demonstrate that truthfulness vectors vary significantly across different semantic contexts through visualization, indicating the necessity for dynamic rather than static intervention approaches.

\item We propose dynamic multimodal activation steering, a training-free method for hallucination mitigation that constructs a semantic-based truthful steering vector database and visual perception steering vector, enabling context-aware interventions during inference by dynamically selecting appropriate steering vectors based on input semantic similarity.

\item We conduct comprehensive experiments on multiple models across discriminative tasks and open-ended generation datasets. The experimental results demonstrate that our method achieves significant improvements: increasing total scores by 94.66 on MME and reducing 20.2\% hallucinations on CHAIR, outperforming existing state-of-the-art methods. These results highlight the effectiveness of our approach in hallucination mitigation.

\end{itemize}

\section{Related Work}

\subsection{Large Vision-Language Models}
Large Vision-Language Models (LVLMs) have recently undergone rapid development, achieving excellent performance in image captioning and VQA tasks \citep{MMsurvey1,MMsurvey2}. They typically consist of a vision encoder, a connection layer, and an LLM. As for the vision encoder, the VIT from CLIP \citep{clip} is commonly used. For the connection layer, some models use simple MLP layers for alignment, such as LLaVA \citep{llava,llava1.5}, Shikra \citep{shikra}, PandaGPT \citep{pandagpt}, etc.; some models use Q-former for alignment, like BLIP2 \citep{blip2}, InstructBLIP \citep{instructblip}, etc.; while others design special architectures. However, these LVLMs suffer from serious hallucination problems, and effectively eliminating hallucinations remains a popular research topic.

\subsection{Hallucination Mitigation for LVLMs}

Recently, numerous approaches have been proposed to mitigate multimodal hallucinations \citep{lvlmhallusurvey,mllmhallusurvey}, addressing this issue across three key stages: training, inference, and post-processing. At the training stage, some research concentrates on constructing better data to train models. For example, LRV \citep{LRV} constructs a high-quality instruction fine-tuning dataset containing balanced positive and negative samples, while other studies introduce reinforcement learning to the multimodal domain to reduce hallucination, such as RLHF-V \citep{rlhf} and RLAIF-V \citep{RLAIF}. These methods typically require carefully constructed training data and consume substantial computational resources during training. Research on mitigating hallucinations at the inference stage often requires no training. VCD \citep{VCD} uses the distribution from noise-added images and the original output distribution to jointly determine the final distribution to mitigate hallucination. 
ICD \citep{ICD} reduces hallucinations by contrasting output distributions between standard and deliberately disturbed instructions. These  contrastive decoding methods often compromise the quality of the generated content \citep{VAF}.
Post-processing approaches correct the generated content from LVLMs to achieve hallucination reduction effects. For instance, LURE \citep{LURE} constructs a dataset to train a hallucination revisor. However, these methods require the construction of a complex pipeline and increase the time required to obtain final outputs. To overcome these limitations, we propose dynamic multimodal activation steering, a training-free approach to mitigate hallucination in LVLMs by dynamically intervening in attention heads during inference time.

\begin{figure}[t!]
  \centering
  \begin{subfigure}[b]{0.34\linewidth}
    \includegraphics[width=\linewidth]{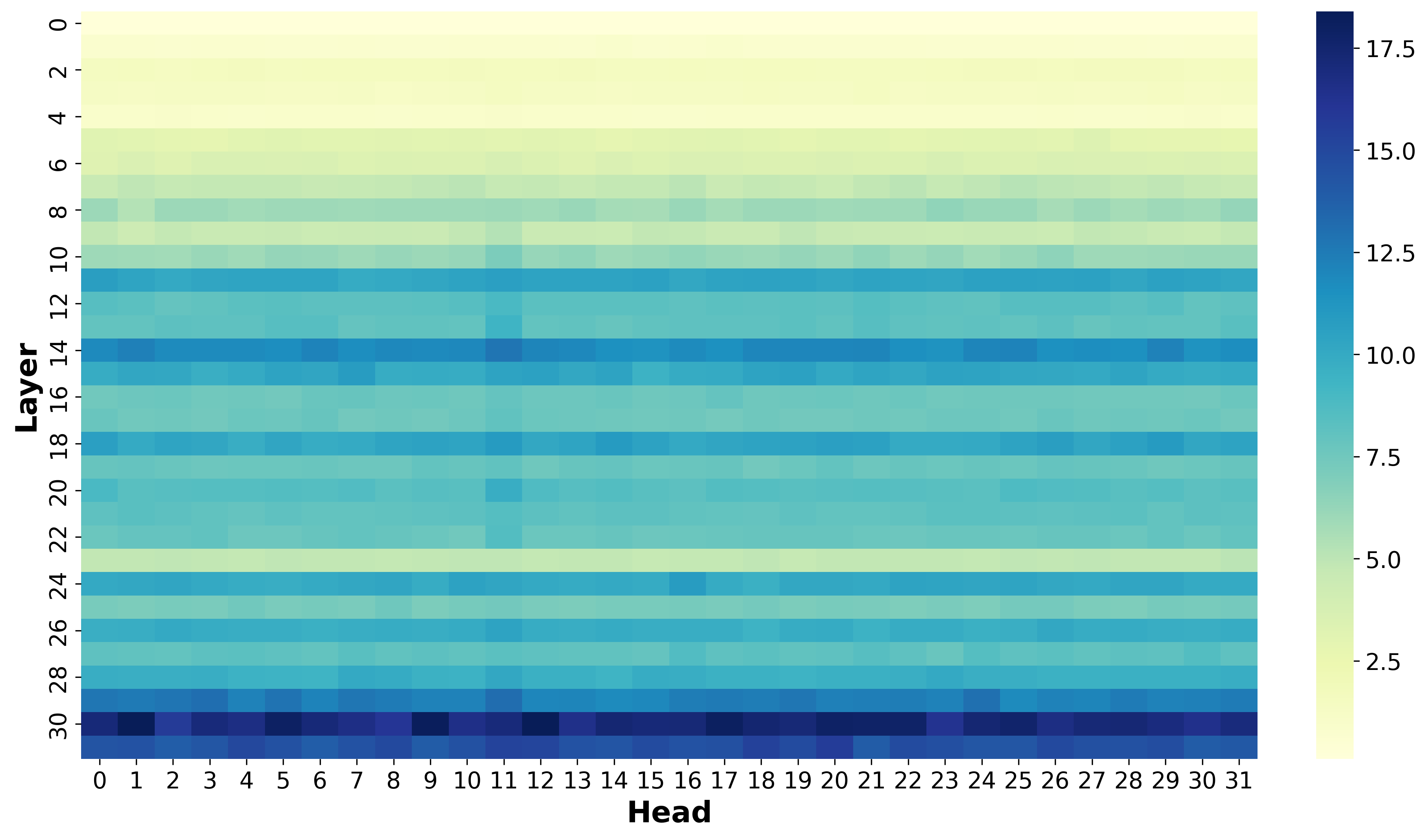}
    \caption{The difference of truthfulness attention head activation.}
    \label{pre1}
  \end{subfigure} \hfill
  \begin{subfigure}[b]{0.34\linewidth}
    \includegraphics[width=\linewidth]{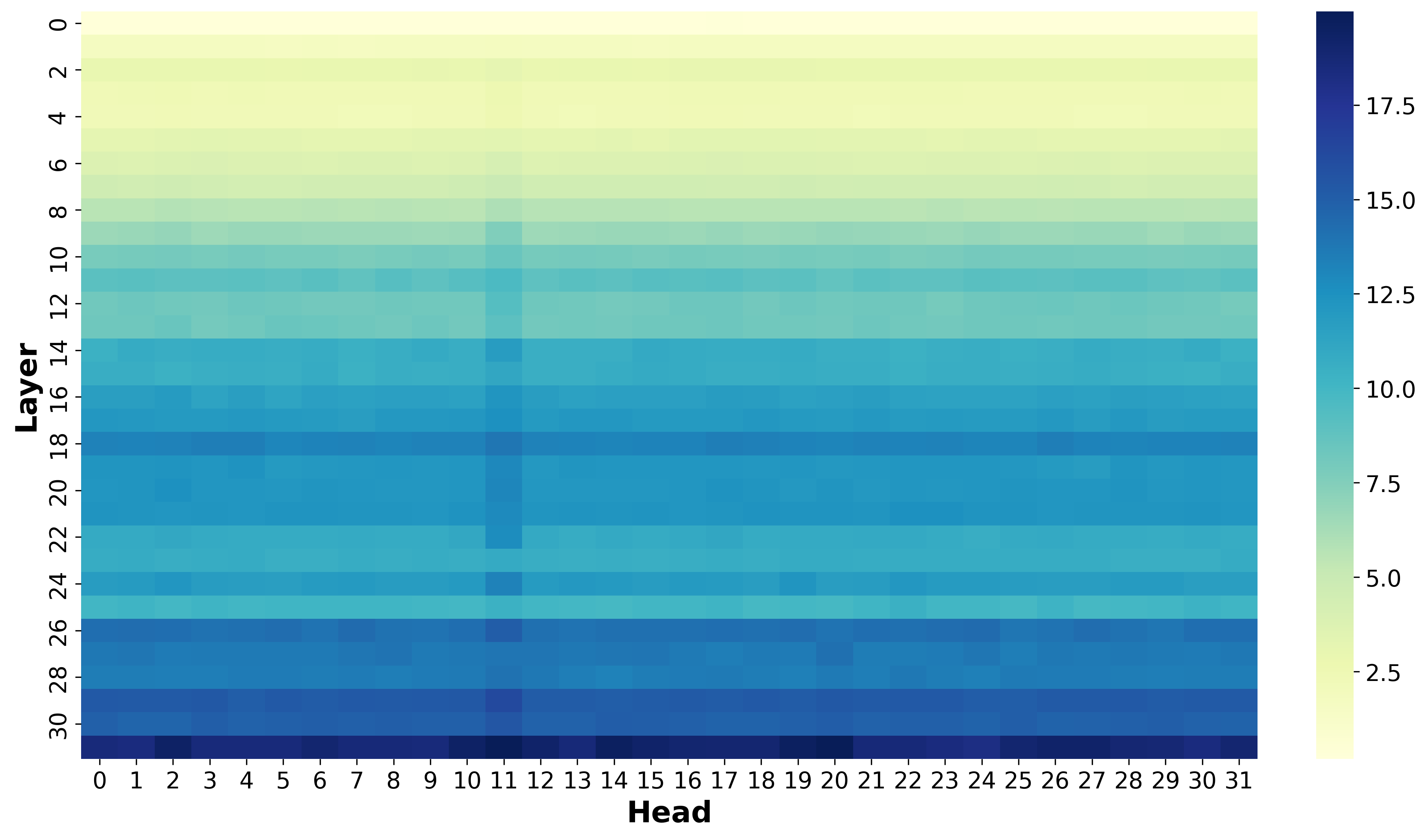}
    \caption{The difference of visual perception attention head activation.}
    \label{pre2}
  \end{subfigure} \hfill
  \begin{subfigure}[b]{0.28\linewidth}
    \includegraphics[width=\linewidth]{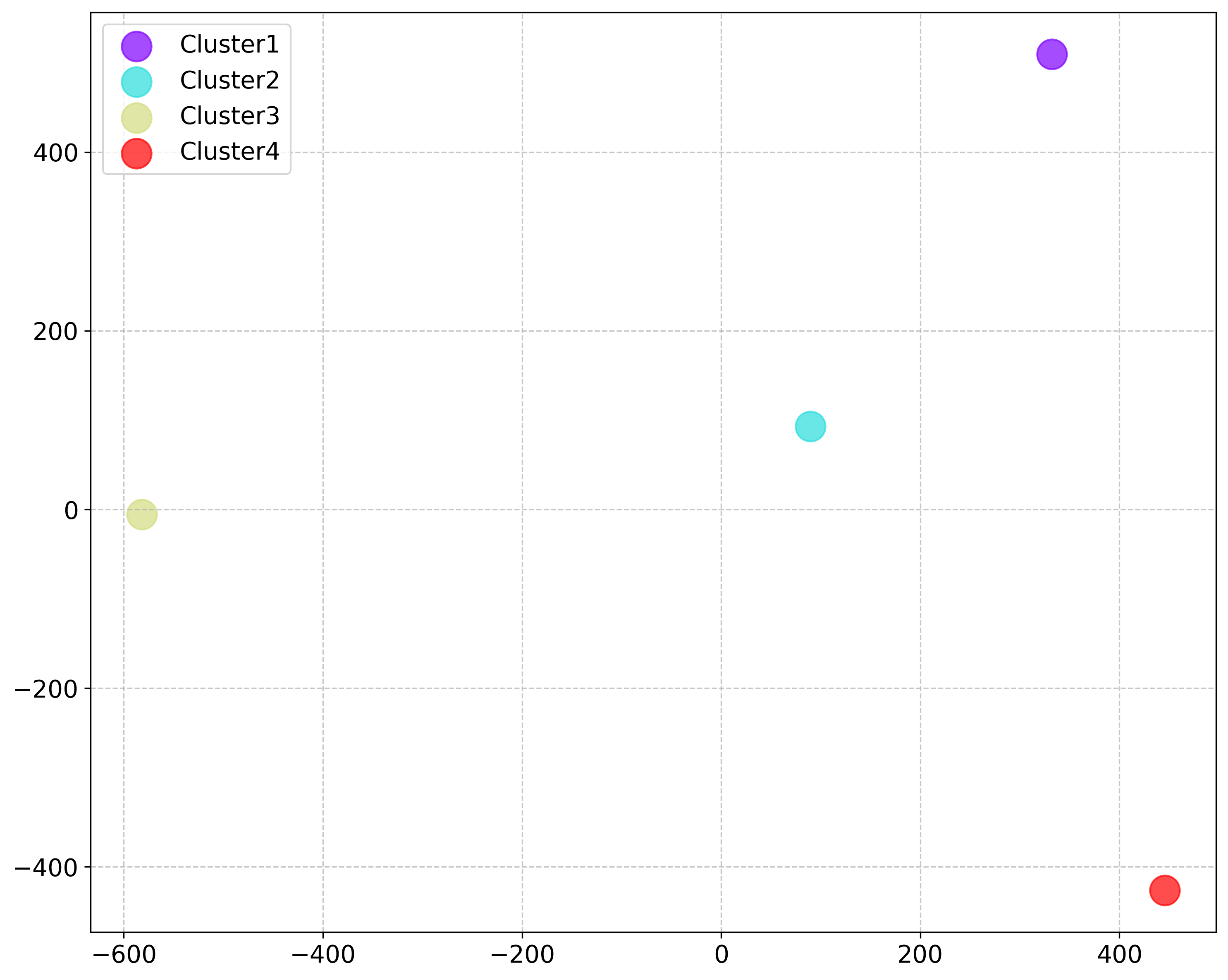}
    \caption{Steering vectors visualization.}
      \label{pre3}
  \end{subfigure}
  \caption{Activation differences in LLaVAv1.5.}
  \label{pre}

\end{figure}

\section{Preliminary Study}
To understand the internal mechanisms underlying multimodal hallucinations, we conduct a systematic analysis of attention patterns in LLaVAv1.5 \citep{llava1.5} across 3,000 samples from the SEED \citep{seed} and AMBER \citep{amber} datasets. Our investigation focuses on identifying which attention heads are most sensitive to truthfulness versus visual perception.

We design two complementary experiments to isolate attention mechanisms responsible for different aspects of multimodal processing. In the first experiment, we examine truthfulness related attention head by contrasting model activations when processing identical visual inputs paired with text prompts either with ground truth or hallucinated answers.  This approach enables us to identify attention heads most relevant to truthfulness, we measure how each head's activation changes between truthful and hallucinated content by computing the difference: truthful activation minus hallucinated activation. In the second experiment, we investigate visual perception related attention head by comparing activations between clean images and their noise-corrupted counterparts, calculating activation differences by subtracting the activation values of non-noisy inputs from those with noise. As shown in Figures \ref{pre1} and \ref{pre2}, the activation patterns differ significantly between these two experiments. For truthfulness (Figure \ref{pre1}), the most active attention heads appear predominantly in layer 30. In contrast, for visual perception (Figure \ref{pre2}), the highest activation differences concentrate in layer 31. 
These distinct activation patterns provide a foundation for our targeted intervention approach that addresses both aspects simultaneously.

Furthermore, we divide the SEED and AMBER datasets into four semantic clusters and compute the activation differences for each cluster. Using t-SNE to visualize these differences in a two-dimensional space (Figure \ref{pre3}), we observe a clear separation between clusters, with each occupying a distinct region in the projection space. This separation indicates that truthfulness direction vectors vary significantly across different semantic contexts. The heterogeneity in these patterns suggests that a static intervention approach would be insufficient, as it cannot account for the semantic-dependent nature of hallucinations. This observation directly motivates our dynamic multimodal activation steering method, which can adaptively select appropriate steering vectors based on the semantic content of the input query.

\section{Method}

\begin{figure*}[thb!]
  \centering
  \includegraphics[width=1\textwidth]{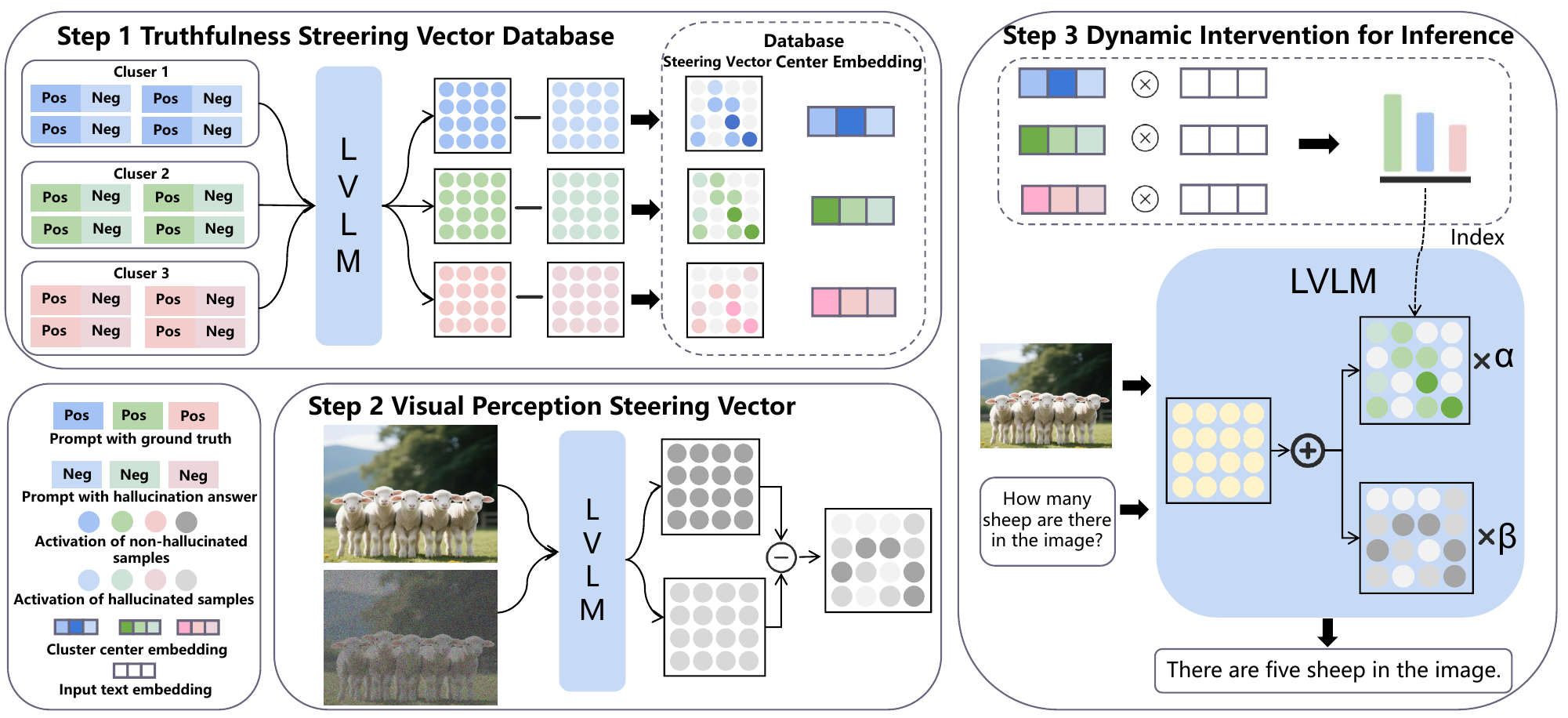}\\
  \caption{Overview of the DMAS framework.}
  \label{framework}
\end{figure*}

In this section, we introduce dynamic multimodal activation steering. As shown in Figure \ref{framework}, the method has three steps: the first step is to establish a dynamic truthfulness steering vector database, the second step is to calculate the steering vector for the model's visual perception attention heads, and the third step is to apply dynamic interventions to different attention heads during inference.

\subsection{Truthfulness Steering Vector Database}
We select the AMBER \citep{amber} and SEED \citep{seed} datasets as our data sources and divide the datasets into 4 clusters based on semantics. The questions in these two datasets are in the form of multiple-choice and discriminative questions, making it easy for us to create hallucinated answers for each sample (for discriminative questions, we change the answer to the opposite; for multiple-choice questions, we randomly select an incorrect option). 
Each cluster $C_i$ comprises the question prompt $T$, visual input $V$, ground truth response $Y_{pos}$, and incorrect response $Y_{neg}$ for every sample.

We input $(V, T+Y_{pos})$ and $(V, T+Y_{neg})$ separately into LVLMs and preserve the attention head activation values of the last token at each layer, denoted as $A_{pos}$ and $A_{neg}$. We define  the  truthfulness steering vector as the activation difference between non-hallucinated outputs and hallucinated outputs within each cluster according to Equation \ref{eq:texthallu}:

\begin{equation}
D_i = \frac{1}{|C_i|} \sum_{j \in C_i} (A_{pos,j} - A_{neg,j})
\label{eq:texthallu}
\end{equation}
$|C_i|$ represents the number of samples in cluster $C_i$, and $j$ indexes the samples within the cluster. Subsequently, we apply principal component analysis (PCA) to $D_i$ to reduce insignificant noise, thereby extracting the principal components that influence truthfulness. The magnitude of $D_i$ effectively quantifies the significance of each attention head in governing this specific model behavior.

Next, we construct a truthful steering vector database where the average embedding representation of questions from each cluster serves as the key, with the corresponding steering vector $D_i$ as the value. During inference, our approach dynamically matches the semantic content of the input question to retrieve the most semantically similar steering vector, enabling context-appropriate interventions. We obtain key embeddings in the database and input text embeddings via sentence transformer.\footnote{\url{https://huggingface.co/sentence-transformers/all-mpnet-base-v2}}

\subsection{Visual Perception Steering Vector}
Given a visual input $V$ and a distorted visual input $V'$ (obtained by adding noise to the image following the forward diffusion process \citep{denoising}), we first input $V$ into an object detector YOLOv11 \citep{yolov11} to obtain objects $O$ present in the image, and insert them into a simple template \textquotesingle The image depicts \{objects\}\textquotesingle~, denoted as ${Y_O}$. Then we randomly select an equal number of objects $O'$ from a predefined object library within the same object category that are not in $O$, inserting them into template, denoted as ${Y_{O'}}$. The prompt $T$ is fixed as \textquotesingle Please describe this image.\textquotesingle~ Next, we obtain the final inputs $(V, T+Y_O)$ and $(V', T+Y_{O'})$, and input these two samples separately into LVLMs, preserving the attention head activation values of the last token at each layer, denoted as $A_v$ and $A_{v'}$ respectively. We define visual perception steering vector as the activation difference between visual input and distorted visual input according to Equation \ref{eq:visualhallu}:
\begin{equation}
D_v = A_v - A_{v'}
\label{eq:visualhallu}
\end{equation}
Similarly, we apply PCA to $D_v$ to reduce noise, thereby extracting the principal components most relevant to visual perception.

\subsection{Dynamic Intervention for Inference}

During the inference phase, for a given text input $T$ and visual input $V$, we dynamically retrieve the most appropriate steering vector by computing semantic similarity between the input and each key in database as shown in 
Equation \ref{eq:dynamic}. 
\begin{equation}
D_f = D_{\hat{i}}, \text{ where } \hat{i} = \underset{i}{\arg\max} \, \text{sim}(E(T), Key_i)
\label{eq:dynamic}
\end{equation}
where $E(T)$ is the embedding representation of the input text, $Key_i$ represents the key embedding for cluster $i$, and $\text{sim}(\cdot,\cdot)$ denotes the cosine similarity function. This process identifies the most relevant truthfulness steering vector for the current input.

To achieve more precise control over model behavior, rather than intervening on all attention heads, we selectively target the most influential heads in both $D_f$ and $D_v$.We define binary mask matrices $M_f$ and $M_v$ as
Equation \ref{eq:mask}:
\begin{equation}
M_{\{f,v\}}^{(l,h)} = \begin{cases} 
1, & \text{if } (l,h) \in \text{TopK}(\mathbf{D}_{\{f,v\}}, K) \\
0, & \text{otherwise}
\end{cases}
\label{eq:mask}
\end{equation}
where $(l,h)$ denotes the $h$-th attention head in the $l$-th layer, $\mathbf{D}$ represents the sum of activation differences for each attention head in $D$ and $\text{TopK}(\mathbf{D}_{\{f,v\}}, K)$ returns the indices of the $K$ largest attention heads in either $D_f$ or $D_v$, representing the most influential attention heads for truthfulness and visual perception respectively.

Building upon the standard attention mechanism, we modify the computation for layers where intervention is applied. Our intervention-enhanced computation is formulated as Equation \ref{eq:attention}:

\begin{equation}
\begin{split}
\mathbf{x}^{(l+1)} &= \mathbf{x}^{(l)} + \text{Concat}_{(0\sim H)} \Big[ \text{Attn}^{(l,h)}(\mathbf{x}^{(l)}) \\
&\quad + \alpha \cdot M_f^{(l,h)} \cdot D_f^{(l,h)} \\
&\quad + \beta \cdot M_v^{(l,h)} \cdot D_v^{(l,h)} \Big] \cdot \mathbf{W}_o^{(l)}
\end{split}
\label{eq:attention}
\end{equation}
where $\mathbf{x}^{(l)}$ represents the hidden states at the $l$-th layer, $H$ is the number of attention heads per layer, $\alpha$ and $\beta$ are hyperparameters controlling the intervention strength for truthfulness and visual perception respectively. The binary masks ensure that interventions are only applied to the most influential attention heads, allowing for precise and targeted steering of the model's behavior.

\begin{table*}[bth!]
  \centering
  \begin{tabular}{cccccccc}
    \toprule
    Model& Method & Existence$\uparrow$ & Count $\uparrow$& Position$\uparrow$& Color$\uparrow$ & Total Scores$\uparrow$  \\
    \midrule
\multirow{9}{*}{LLaVAv1.5}     & Regular & 175.67 & 124.67 & 114.00 & 151.00 & 565.33 \\
                              & VCD & 184.66 & 138.33 & 128.67 & 153.00 & 604.66 \\
                              & OPERA & 180.67 &133.33 &123.33 &155.00 &592.33\\
                              & VAF & \textbf{195.00}  & 158.33 & 128.33 & 155.00 & 636.67 \\
                              & DECO & 185.00	&153.33	&118.33	&155.00	&611.66\\
                              & DAMO & \textbf{195.00}	&150.00	&\textbf{143.33}	&165.00	&653.33\\
                              & AGLA & \textbf{195.00}  & 153.89 & 129.44 & 161.67 & 640.00 \\
                              & ICT & 190.00  & \textbf{160.43} & 128.67 & 170.00 & 649.10 \\
                             &  Ours & \textbf{195.00}   & 158.33 & 133.33 & \textbf{173.33} & \textbf{659.99}\\
                              % & $\Delta$ & \textcolor{red}{$\uparrow$19.33} & \textcolor{red}{$\uparrow$33.66} & \textcolor{red}{$\uparrow$19.33} & \textcolor{red}{$\uparrow$22.33} & \textcolor{red}{$\uparrow$94.66}\\
\midrule
\multirow{5}{*}{QwenVL}  & Regular & 155.00 & 127.67 & 131.67 & 173.00 & 587.33 \\
                         & VCD & 156.00 & 131.00 & 128.00 & 181.67 & 596.67 \\
                         & VAF & 165.00 & \textbf{155.00} & \textbf{133.33} & 175.00 & 628.33 \\
                         & ICT & \textbf{180.00} &145.00 & 108.33 & 173.33 & 606.66 \\
                         &  Ours & 170.00 & 145.00 & \textbf{133.33} & \textbf{185.00} & \textbf{633.33} \\
    \bottomrule
  \end{tabular}
  \caption{Results on MME. The best results are shown in bold.}
  \label{tab:MME}
\end{table*}

\section{Experimental Setup}

\subsection{Datasets and Evaluation Metrics}
To comprehensively evaluate our proposed approach, we test our method on discriminative tasks, including MME \citep{MME} and POPE \citep{POPE}, as well as on open-ended generation tasks using CHAIR \citep{Chair}.

\paragraph{MME} \citep{MME} is a comprehensive evaluation benchmark for LVLMs, comprising 14 subtasks. For questions in this dataset, models are required to respond with either \textquotesingle yes\textquotesingle~or \textquotesingle no\textquotesingle. Following \citet{woodpecker} and \citet{VCD}, we select \textquotesingle existence\textquotesingle, \textquotesingle count\textquotesingle, \textquotesingle position\textquotesingle, and \textquotesingle attribute\textquotesingle~as the hallucination test sets. Consistent with \citet{MME}, we adopt the sum of accuracy and accuracy+ as the evaluation metrics.

\paragraph{POPE} \citep{POPE}  is a benchmark designed specifically to evaluate object hallucination. The benchmark features three sampling strategies of varying difficulty levels: random (randomly sampling nonexistent objects), popular (selecting frequently appearing objects), and adversarial (selecting objects that frequently co-occur with objects present in the image). We report Accuracy, Precision, Recall, F1 Score as the evaluation metrics.

\paragraph{CHAIR} \citep{Chair} is an open-ended generation task. This benchmark comprises 500 images sourced from MSCOCO \citep{coco}, where LVLMs are required to generate captions for the images, followed by evaluation of hallucinations present in these captions at sentence level CHAIR$_{S}$ and image level CHAIR$_{I}$.

\subsection{Baselines and Implementation Details}
We validate the effectiveness on mainstream LVLMs: LLaVAv1.5 7B \citep{llava1.5} and QwenVL 7B \citep{qwenvl}, and compare DMAS with state-of-the-art methods: ICT \citep{ICT}, AGLA \citep{AGLA}, VAF \citep{VAF}, VTI \citep{VTI}, DECO \citep{deco}, DAMO \citep{damo}, VCD \citep{VCD}, and OPERA \citep{opera}.

Our method has three key parameters: $\alpha$, $\beta$, and $K$. $\alpha$ and $\beta$ respectively regulate the intensity of interventions for truthfulness and visual perception, while $K$ refers to the intervention on the top $K$ most active attention heads. We set the range of $\alpha$ and $\beta$ to \{0.5, 1, 2, 3, 4, 5, 6, 7, 8, 9, 10\}, and the range of $K$ to \{32, 64, 128, 256, 512, 1024\}, and employ grid search to determine the parameters. In our experiments, we set the temperature to 0 and top\_p to 1. All experiments are conducted on NVIDIA RTX 4090(48GB) GPUs.

\section{Experiment}
\subsection{Results on MME}

The results on the MME \citep{MME} dataset are presented in Table \ref{tab:MME}. Our method demonstrates significant improvements of 94.66 and 46 points compared to the baseline models LLaVAv1.5 and QwenVL, respectively. On the LLaVAv1.5 model, our approach outperforms the existing state-of-the-art method ICT \citep{ICT} by 10.89 points, while on QwenVL, it surpasses the current state-of-the-art method VAF \citep{VAF} by 5 points. Across all subtasks, we observe notable improvements over regular baselines, which can be attributed to our dynamic  intervention mechanism that retrieves the most semantically similar steering vector for each query.

\subsection{Results on POPE}

\begin{table*}[thb!]
  \centering
  \begin{tabular}{ccccccc}
    \toprule
    Dataset& Setting & Method & Accuracy $\uparrow$ & Precision & Recall & F1 Score $\uparrow$ \\
    \midrule
\multirow{15}{*}{MSCOCO} & \multirow{8}{*}{LLaVAv1.5} & Regular & 81.38	& 88.04	& 72.78	&79.65 \\
                              &                    & VCD & 84.33	&85.93	&83.28	&84.52 \\
                               &                    & OPERA &84.21	&88.23	&79.79	&83.72 \\
                               &                    & VAF &86.90	&89.43	&83.77	&86.47 \\
                               &                    & AGLA &85.82	&93.78	&76.83	&84.44 \\
                               &                    & VTI &86.48	&90.11	&82.09	&85.90 \\
                              &                    & ICT & \textbf{87.35} & - & - & \textbf{87.12} \\
                             &                    &  Ours & \underline{86.81} & 87.23 & 86.57 & \underline{86.79} \\
\cmidrule{2-7}
                              & \multirow{7}{*}{QwenVL} & Regular & 83.71	&93.30 	&72.69	&81.70 \\
                              &                    & VCD & 86.67	&90.66	&81.94	&83.04\\
                              &                    & OPERA &84.26	&94.40 	&73.52	&82.65 \\
                              &                    & AGLA &83.9	&96.20 	&70.62	&81.44 \\
                              &                    & VTI &85.18	&91.31	&78.18	&84.08 \\
                              &                    & ICT & \underline{87.53} & - & - & \underline{86.98} \\
                             &                    &  Ours & \textbf{87.63} &87.92	&87.3	&\textbf{87.65} \\ 
\midrule
\multirow{13}{*}{GQA}  & \multirow{7}{*}{LLaVAv1.5} & Regular & 78.33 	&79.33	&79.13	&79.13\\
                              &                    & VCD & 81.16	&77.31	&89.08	&82.67 \\
                              &                    & OPERA & 80.80	& -&-&83.24 \\
                              &                    & VAF & 83.67	&81.50	&88.00	&84.50 \\
                              &                    & AGLA &\underline{84.41}	&84.63	&84.67	&84.55 \\
                              &                    & ICT & \textbf{85.27} & - & - & \underline{85.50} \\
                             &                    &  Ours & \textbf{85.27}	&83.86	&87.51	&\textbf{85.63} \\ 
\cmidrule{2-7}                              
                              & \multirow{6}{*}{QwenVL} & Regular & 77.47	&81.54	&71.37	&76.06 \\
                              &                    & VCD & 82.48	&81.73	&83.93	&82.77 \\
                              &                    & OPERA & 82.74 & -&-&82.68 \\
                              &                    & AGLA &81.14	&86.87	&73.53	&79.63 \\
                              &                    & ICT & \underline{83.28} & - & - & \underline{83.26} \\
                             &                    &  Ours & \textbf{84.40}	&85.19	&83.53	&\textbf{84.32}\\ 
    \bottomrule
  \end{tabular}
  \caption{Results on POPE. Best results are in bold, and second-best values are underlined.}
  \label{tab:POPE}
\end{table*}

The experimental results of POPE \citep{POPE} are shown in Table \ref{tab:POPE}. We conduct experiments on MSCOCO \citep{coco} and GQA \citep{gqa} under random, popular, and adversarial settings. Table \ref{tab:POPE} presents the average results across these three settings, with detailed experimental results provided in the Appendix. Our method improves LLaVAv1.5's performance on MSCOCO by 5.43\% in accuracy and 7.14\% in F1 score, while for QwenVL, it achieved improvements of 3.92\% in accuracy and 5.95\% in F1 score. On GQA, our method enhances LLaVAv1.5 by 6.94\% in accuracy and 6.5\% in F1 score, and improves QwenVL by 6.93\% in accuracy and 8.26\% in F1 score. Compared to existing methods, our approach achieves best results in most cases, demonstrating its significant effectiveness in mitigating object hallucination. Notably, while the ICT \citep{ICT} method applies noise to objects in images to increase the LVLMs' attention to these objects, our method achieves superior performance in most cases without such specialized design elements.

\subsection{Results on CHAIR}

We evaluate our method on open-ended generation tasks, with experimental results on CHAIR \citep{Chair} presented in Table \ref{tab:CHAIR_original}. Our method reduces hallucinations by 20.2 at the sentence level (CHAIR$_S$) and by 3.8 at the image level (CHAIR$_I$). Compared to existing methods, our approach reduces sentence-level hallucinations by 5 points over the state-of-the-art method VTI \citep{VTI}, and matching VTI's performance on image-level hallucinations. In summary, our method achieves significant improvements in hallucination mitigation on both discriminative tasks and open-ended generation tasks.

\begin{table}[thb!]
  \centering
  \begin{minipage}[t]{0.3\linewidth}
    \centering
    \begin{tabular}{ccc}
      \toprule
      Method & $CHAIR_S$$\downarrow$ & $CHAIR_I$$\downarrow$  \\
      \midrule
      Regular & 51.0 & 15.2  \\
      VCD & 51.0 & 14.9  \\
      OPERA & 47.0 & 14.6  \\
      DECO & 37.8	&\textbf{11.1} \\
      AGLA   & 43.0 & 14.1 \\
      VTI & 35.8 & \textbf{11.1} \\
      Ours & \textbf{30.8} & 11.4 \\
      \bottomrule
    \end{tabular}
    \caption{Results on CHAIR.}  
    \label{tab:CHAIR_original}  
  \end{minipage}
  \hfill  
  \begin{minipage}[t]{0.6\linewidth}
    \centering
    \begin{tabular}{ccccc}
      \toprule
      \multirow{2}{*}{Method} & \multicolumn{2}{c}{CHAIR} & \multicolumn{2}{c}{POPE} \\
      \cmidrule{2-3} \cmidrule{4-5}
       & C$_S$$\downarrow$ & C$_I$$\downarrow$ & Acc$\uparrow$ & F1$\uparrow$ \\
      \midrule
      Ours & \textbf{30.8} & \textbf{11.4} & \textbf{81.70} & \textbf{82.47} \\
      w/o visual vector & 34.2 & 11.7 & 81.67 & 82.42 \\
      w/o truthfulness vector & 42.4 & 13.2 & 81.40 & 82.01 \\
      w/o both & 51.0 & 15.2 & 75.08 & 76.06 \\
      \bottomrule
    \end{tabular}
    \caption{Ablation studies on CHAIR and POPE. C$_S$ represents CHAIR$_S$, C$_I$ represents CHAIR$_I$.}  
    \label{tab:ablation} 
  \end{minipage}
\end{table}

\subsection{Further Analysis}

\subsubsection{Ablation Studies}

To demonstrate the effectiveness of using both truthfulness steering vector and visual perception steering vector, we compare the results when utilizing only one intervention at a time. We conduct experiments on LLaVAv1.5 using the CHAIR and POPE. As shown in Table \ref{tab:ablation}, \textquotesingle w/o visual vector\textquotesingle~ indicates intervention with only the truthfulness steering vector, while \textquotesingle w/o truthfulness vector\textquotesingle~ indicates intervention with only the visual perception steering vector. We observe that even when using only one intervention method, there is a notable improvement compared to the Regular baseline (w/o both). Furthermore, each intervention method exhibits hallucination mitigation effects on both discriminative and generation tasks. The optimal results are achieved when the two interventions are combined.

\begin{figure*}[]
  \centering
  \begin{subfigure}[b]{0.32\linewidth}
    \includegraphics[width=\linewidth]{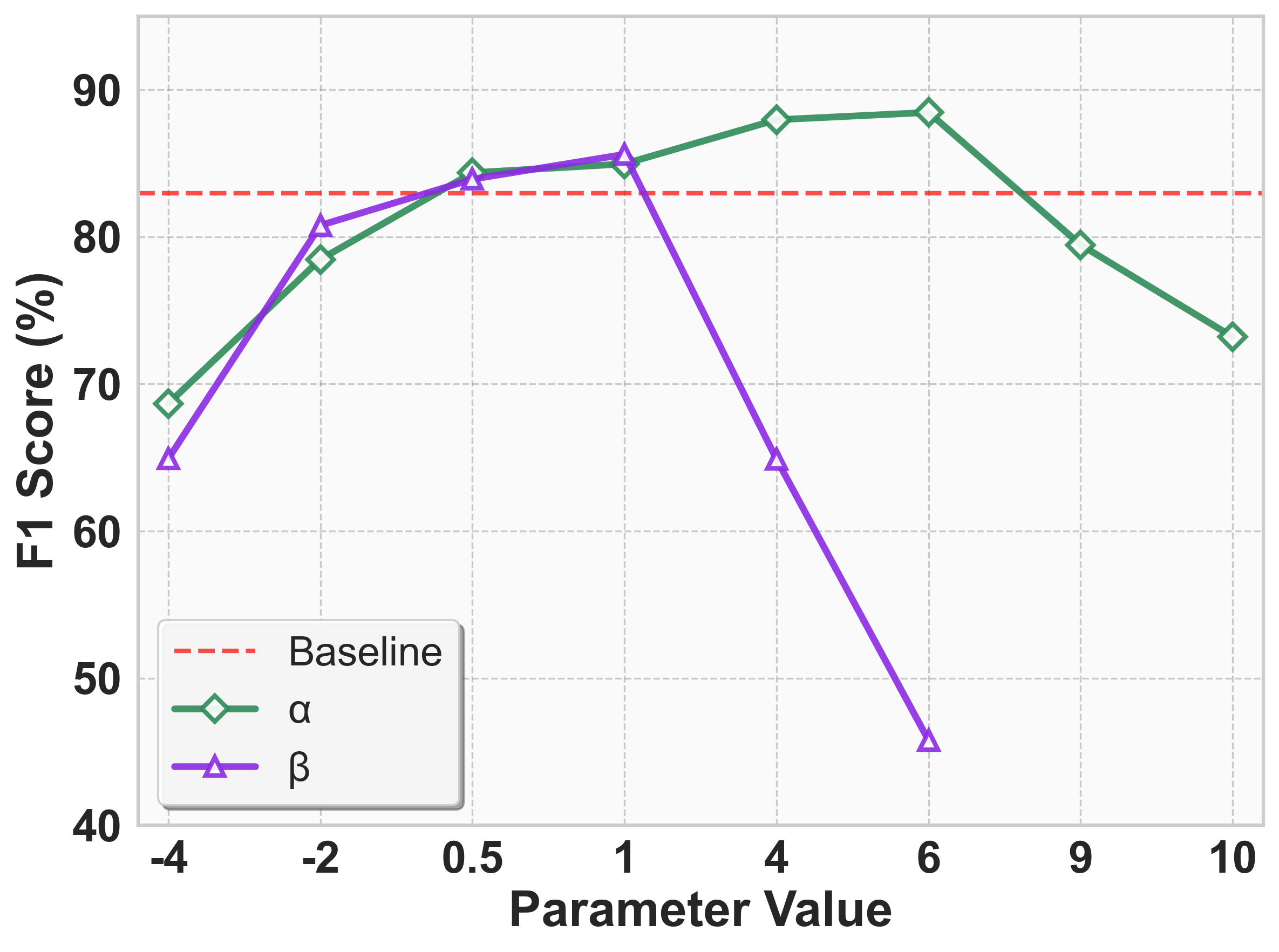}
    \caption{Impact of $\alpha$ and $\beta$.}
    \label{para1}
  \end{subfigure}
  \hfill
  \begin{subfigure}[b]{0.32\linewidth}
    \includegraphics[width=\linewidth]{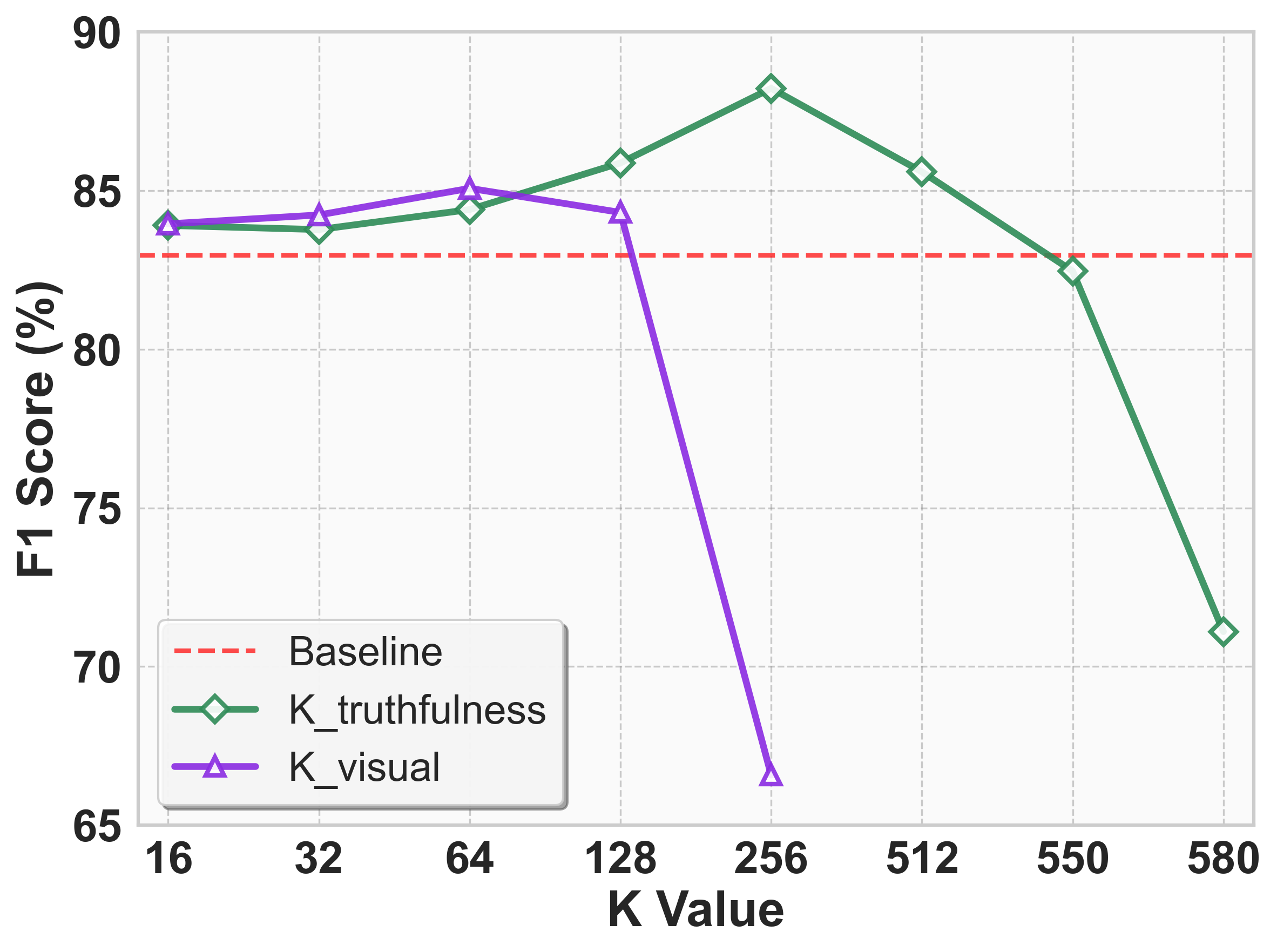}
    \caption{Impact of $K$.}
    \label{para2}
  \end{subfigure}
  \hfill
  \begin{subfigure}[b]{0.32\linewidth}
    \includegraphics[width=\linewidth]{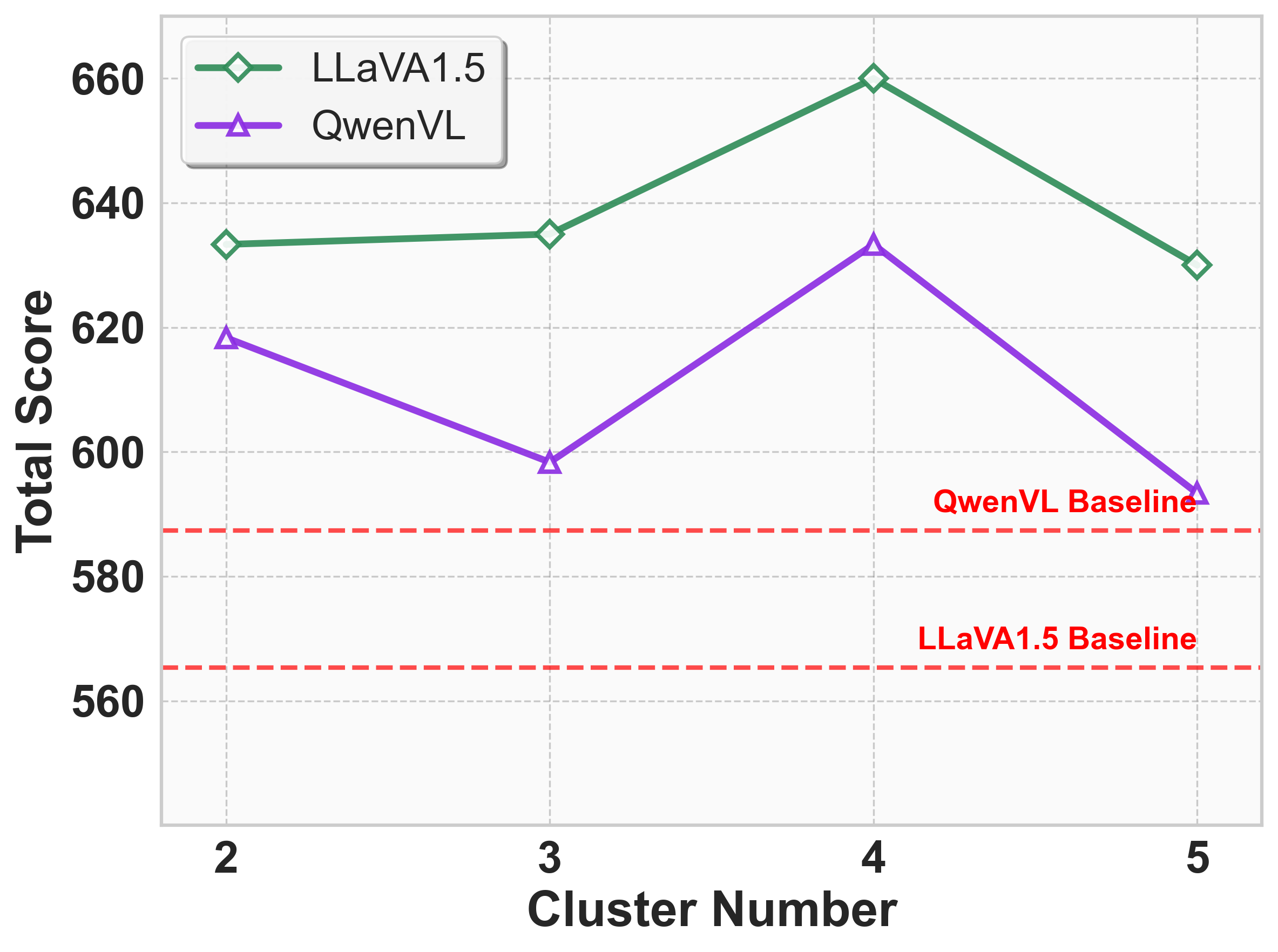}
    \caption{Impact of Clusters.}
    \label{para3}
  \end{subfigure}
  \caption{Impact of key hyperparameters\textcolor{blue}{.}}
  \label{para}
\end{figure*}

\subsubsection{Effect of Dynamic Intervention}

\begin{wrapfigure}{r}{0.5\textwidth}  
  \centering  
  \includegraphics[width=\linewidth]{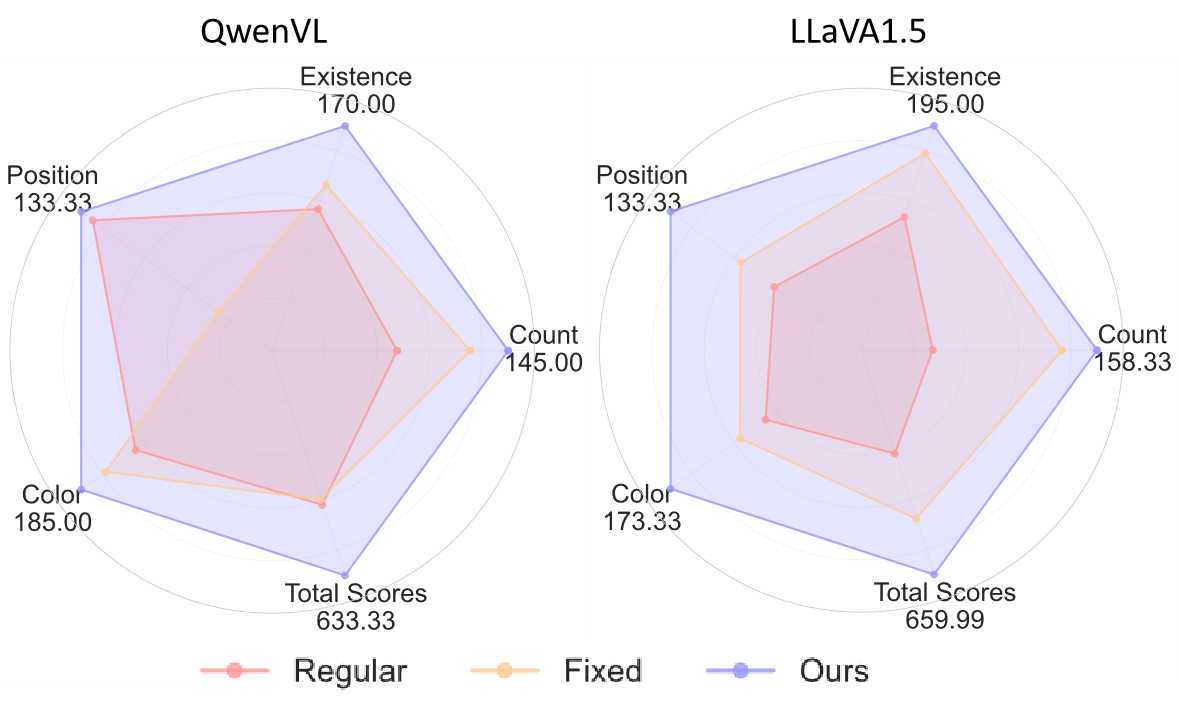} 
  \caption{Effect of dynamic intervention.}  
  \label{dynamic}  
\end{wrapfigure}

To validate our designed strategy of dynamically invoking the truthfulness steering vector based on semantics, we compare our method with combining all truthfulness steering vectors into a single fixed steering vector for intervention. We conduct experiments on QwenVL and LLaVAv1.5 using the MME, with results shown in Figure \ref{dynamic}. We observe that dynamically invoking steering vectors based on semantics achieves optimal performance across all subtasks. When using a fixed steering vector, the improvement is smaller than with our method, and it even underperforms the original model on the Position subtask of QwenVL, which demonstrates the necessity of our designed dynamic invocation strategy.

\subsubsection{Impact of Hyperparameters}

In this section, we investigate the impact of key parameters $\alpha$, $\beta$, and $K$ on the experimental results. Here, $\alpha$ and $\beta$ control the intervention strength, while $K$ denotes the number of attention heads receiving intervention. Experiments conducted on QwenVL using the POPE GQA random subset are shown in Figure \ref{para}. Figure \ref{para1} illustrates the relationship between F1 score and parameters $\alpha$ and $\beta$. When $\alpha$ and $\beta$ are negative, we observe a decrease in F1 score, which effectively represents intervention in the opposite direction, pushing activations toward hallucination. As $\alpha$ and $\beta$ increase, F1 score exhibits an upward trend; however, when $\alpha$ and $\beta$ become excessively large, F1 score shows a precipitous decline, indicating that the model's fundamental capabilities become impaired. Figure \ref{para2} shows how F1 varies with the number of intervened attention heads, revealing similar patterns for both truthfulness and visual perception attention heads. Few intervened heads produce minimal impact with no significant F1 improvement. As intervention extends to more heads, F1 score increases, but excessive intervention causes a dramatic decline in F1, indicating degradation of model performance.

\subsubsection{Impact of Clusters}
In this section, we investigate the impact of cluster quantity on the performance of our proposed method. We vary the number of clusters across \{2, 3, 4, 5\} and conduct experiments on both QwenVL and LLaVAv1.5 using the MME benchmark. The experimental results are presented in Figure \ref{para3}. We observe that both LLaVAv1.5 and QwenVL achieve optimal performance when the number of clusters is set to 4. When the cluster count is insufficient, the semantic granularity becomes too coarse for effective representation.

\subsubsection{Analysis of Generality}
To verify the generalizability of our method, we tested our approach on ScienceQA \citep{scienceqa} which is subject-based VQA dataset  and ViQuAE \citep{Viquae} which is a knowledge-based VQA dataset. The accuracy are shown in Table \ref{tab:gen}. Our method also achieved significant improvements on these datasets. These datasets are completely different from the dataset types we used to construct the steering vector, which demonstrates the generalizability of our method.

\begin{table}[thb!]
\centering
% \small
\setlength{\tabcolsep}{3pt} 
\begin{tabular}{ccccc}
\toprule
\multirow{2}{*}{Method} & \multicolumn{2}{c}{ScienceQA} & \multicolumn{2}{c}{ViQuAE} \\
\cmidrule{2-3} \cmidrule{4-5}
 & LLaVAv1.5 & QwenVL & LLaVAv1.5 & QwenVL  \\
\midrule
Regular  & 52.75 & 46.41 & 43.38 & 50.09 \\
VTI & 51.46 & - & 42.29 & - \\
ICT & 52.95 & - & 42.47 & - \\
Ours & \textbf{62.27} & \textbf{48.04} & \textbf{56.00} & \textbf{54.08} \\
\bottomrule
\end{tabular}
\caption{Generality on ScienceQA and ViQuAE.}
\label{tab:gen}
\end{table}

\section{Conclusion}
This paper proposes dynamic multimodal activation steering, a training-free approach to mitigate hallucination in LVLMs by dynamically intervening in attention head activations.  The experiments on multiple benchmarks demonstrate the effectiveness of our method, with LLaVAv1.5 achieving a remarkable 94.66-point improvement on MME and reducing 20.2\% hallucinations on CHAIR, outperforming existing SOTA methods. We compare the experimental performance of our proposed semantic-dynamic strategy for steering vector selection against  fixed steering vector approaches, demonstrating the effectiveness and necessity of semantic-dynamic steering vector selection.

\subsubsection*{Acknowledgments}
This research is funded by the National Nature Science Foundation of China (No. 62477010, No.62577022 and No.62307028), the Natural Science Foundation of Shanghai (No. 23ZR1441800), Shanghai Science and Technology Innovation Action Plan (No. 24YF2710100 and No.23YF1426100),Shanghai Qiji Zhifeng Co., Ltd. (2025-GZL-RGZN-01001) and the opening funding of the State Key Laboratory of DisasterReduction in Civil Engineering (Grant No. SLDRCE24-03).

% \bibliography{iclr2026_conference}

\begin{thebibliography}{57}
\providecommand{\natexlab}[1]{#1}
\providecommand{\url}[1]{\texttt{#1}}
\expandafter\ifx\csname urlstyle\endcsname\relax
  \providecommand{\doi}[1]{doi: #1}\else
  \providecommand{\doi}{doi: \begingroup \urlstyle{rm}\Url}\fi

\bibitem[An et~al.(2025)An, Tian, Leng, Nie, Lin, Wang, Chen, Zhang, and Lu]{AGLA}
Wenbin An, Feng Tian, Sicong Leng, Jiahao Nie, Haonan Lin, QianYing Wang, Ping Chen, Xiaoqin Zhang, and Shijian Lu.
\newblock Mitigating object hallucinations in large vision-language models with assembly of global and local attention.
\newblock In \emph{Proceedings of the Computer Vision and Pattern Recognition Conference}, pp.\  29915--29926, 2025.

\bibitem[Bai et~al.(2023)Bai, Bai, Yang, Wang, Tan, Wang, Lin, Zhou, and Zhou]{qwenvl}
Jinze Bai, Shuai Bai, Shusheng Yang, Shijie Wang, Sinan Tan, Peng Wang, Junyang Lin, Chang Zhou, and Jingren Zhou.
\newblock Qwen-vl: A versatile vision-language model for understanding, localization, text reading, and beyond, 2023.
\newblock URL \url{https://arxiv.org/abs/2308.12966}.

\bibitem[Bai et~al.(2025)Bai, Chen, Liu, Wang, Ge, Song, Dang, Wang, Wang, Tang, et~al.]{qwen25vl}
Shuai Bai, Keqin Chen, Xuejing Liu, Jialin Wang, Wenbin Ge, Sibo Song, Kai Dang, Peng Wang, Shijie Wang, Jun Tang, et~al.
\newblock Qwen2. 5-vl technical report.
\newblock \emph{arXiv preprint arXiv:2502.13923}, 2025.

\bibitem[Bai et~al.(2024)Bai, Wang, Xiao, He, Han, Zhang, and Shou]{mllmhallusurvey}
Zechen Bai, Pichao Wang, Tianjun Xiao, Tong He, Zongbo Han, Zheng Zhang, and Mike~Zheng Shou.
\newblock Hallucination of multimodal large language models: A survey.
\newblock \emph{arXiv preprint arXiv:2404.18930}, 2024.

\bibitem[Chen et~al.(2025{\natexlab{a}})Chen, Zhang, Huang, Niu, Zhang, Wen, and Hu]{ICT}
Junzhe Chen, Tianshu Zhang, Shiyu Huang, Yuwei Niu, Linfeng Zhang, Lijie Wen, and Xuming Hu.
\newblock Ict: Image-object cross-level trusted intervention for mitigating object hallucination in large vision-language models.
\newblock In \emph{Proceedings of the Computer Vision and Pattern Recognition Conference}, pp.\  4209--4221, 2025{\natexlab{a}}.

\bibitem[Chen et~al.(2025{\natexlab{b}})Chen, Chen, Zhou, Tao, Ding, Xie, Xie, Li, and Feng]{hallucination2}
Kedi Chen, Qin Chen, Jie Zhou, Xinqi Tao, Bowen Ding, Jingwen Xie, Mingchen Xie, Peilong Li, and Zheng Feng.
\newblock Enhancing uncertainty modeling with semantic graph for hallucination detection.
\newblock In \emph{Proceedings of the AAAI Conference on Artificial Intelligence}, volume~39, pp.\  23586--23594, 2025{\natexlab{b}}.

\bibitem[Chen et~al.(2023)Chen, Zhang, Zeng, Zhang, Zhu, and Zhao]{shikra}
Keqin Chen, Zhao Zhang, Weili Zeng, Richong Zhang, Feng Zhu, and Rui Zhao.
\newblock Shikra: Unleashing multimodal llm's referential dialogue magic.
\newblock \emph{arXiv preprint arXiv:2306.15195}, 2023.

\bibitem[Chen et~al.(2024)Chen, Wang, Cao, Liu, Gao, Cui, Zhu, Ye, Tian, Liu, et~al.]{internVL}
Zhe Chen, Weiyun Wang, Yue Cao, Yangzhou Liu, Zhangwei Gao, Erfei Cui, Jinguo Zhu, Shenglong Ye, Hao Tian, Zhaoyang Liu, et~al.
\newblock Expanding performance boundaries of open-source multimodal models with model, data, and test-time scaling.
\newblock \emph{arXiv preprint arXiv:2412.05271}, 2024.

\bibitem[Cui et~al.(2024)Cui, Ma, Cao, Ye, Zhou, Liang, Chen, Lu, Yang, Liao, et~al.]{autonomousdriving}
Can Cui, Yunsheng Ma, Xu~Cao, Wenqian Ye, Yang Zhou, Kaizhao Liang, Jintai Chen, Juanwu Lu, Zichong Yang, Kuei-Da Liao, et~al.
\newblock A survey on multimodal large language models for autonomous driving.
\newblock In \emph{Proceedings of the IEEE/CVF Winter Conference on Applications of Computer Vision}, pp.\  958--979, 2024.

\bibitem[Dai et~al.(2023)Dai, Li, Li, Tiong, Zhao, Wang, Li, Fung, and Hoi]{instructblip}
Wenliang Dai, Junnan Li, Dongxu Li, Anthony Meng~Huat Tiong, Junqi Zhao, Weisheng Wang, Boyang Li, Pascale Fung, and Steven Hoi.
\newblock Instructblip: Towards general-purpose vision-language models with instruction tuning, 2023.
\newblock URL \url{https://arxiv.org/abs/2305.06500}.

\bibitem[Fu et~al.(2023)Fu, Chen, Shen, Qin, Zhang, Lin, Qiu, Lin, Yang, Zheng, Li, Sun, and Ji]{MME}
Chaoyou Fu, Peixian Chen, Yunhang Shen, Yulei Qin, Mengdan Zhang, Xu~Lin, Zhenyu Qiu, Wei Lin, Jinrui Yang, Xiawu Zheng, Ke~Li, Xing Sun, and Rongrong Ji.
\newblock {MME:} {A} comprehensive evaluation benchmark for multimodal large language models.
\newblock \emph{CoRR}, abs/2306.13394, 2023.
\newblock \doi{10.48550/ARXIV.2306.13394}.
\newblock URL \url{https://doi.org/10.48550/arXiv.2306.13394}.

\bibitem[Guan et~al.(2024)Guan, Liu, Wu, Xian, Li, Liu, Wang, Chen, Huang, Yacoob, et~al.]{hallusionbench}
Tianrui Guan, Fuxiao Liu, Xiyang Wu, Ruiqi Xian, Zongxia Li, Xiaoyu Liu, Xijun Wang, Lichang Chen, Furong Huang, Yaser Yacoob, et~al.
\newblock Hallusionbench: an advanced diagnostic suite for entangled language hallucination and visual illusion in large vision-language models.
\newblock In \emph{Proceedings of the IEEE/CVF Conference on Computer Vision and Pattern Recognition}, pp.\  14375--14385, 2024.

\bibitem[Ho et~al.(2020)Ho, Jain, and Abbeel]{denoising}
Jonathan Ho, Ajay Jain, and Pieter Abbeel.
\newblock Denoising diffusion probabilistic models.
\newblock \emph{Advances in neural information processing systems}, 33:\penalty0 6840--6851, 2020.

\bibitem[Huang et~al.(2025)Huang, Yu, Ma, Zhong, Feng, Wang, Chen, Peng, Feng, Qin, et~al.]{huang2025survey}
Lei Huang, Weijiang Yu, Weitao Ma, Weihong Zhong, Zhangyin Feng, Haotian Wang, Qianglong Chen, Weihua Peng, Xiaocheng Feng, Bing Qin, et~al.
\newblock A survey on hallucination in large language models: Principles, taxonomy, challenges, and open questions.
\newblock \emph{ACM Transactions on Information Systems}, 43\penalty0 (2):\penalty0 1--55, 2025.

\bibitem[Huang et~al.(2024)Huang, Dong, Zhang, Wang, He, Wang, Lin, Zhang, and Yu]{opera}
Qidong Huang, Xiaoyi Dong, Pan Zhang, Bin Wang, Conghui He, Jiaqi Wang, Dahua Lin, Weiming Zhang, and Nenghai Yu.
\newblock Opera: Alleviating hallucination in multi-modal large language models via over-trust penalty and retrospection-allocation.
\newblock In \emph{Proceedings of the IEEE/CVF Conference on Computer Vision and Pattern Recognition}, pp.\  13418--13427, 2024.

\bibitem[Hudson \& Manning(2019)Hudson and Manning]{gqa}
Drew~A Hudson and Christopher~D Manning.
\newblock Gqa: A new dataset for real-world visual reasoning and compositional question answering.
\newblock In \emph{Proceedings of the IEEE/CVF conference on computer vision and pattern recognition}, pp.\  6700--6709, 2019.

\bibitem[Jiang et~al.()Jiang, He, Zeng, Wei, Ku, Liu, and Chen]{mantis}
Dongfu Jiang, Xuan He, Huaye Zeng, Cong Wei, Max Ku, Qian Liu, and Wenhu Chen.
\newblock Mantis: Interleaved multi-image instruction tuning.
\newblock \emph{Transactions on Machine Learning Research}.

\bibitem[Jin et~al.(2024)Jin, Li, Liu, Gu, Wu, Jiang, He, Zhao, Tan, Gan, et~al.]{MMsurvey2}
Yizhang Jin, Jian Li, Yexin Liu, Tianjun Gu, Kai Wu, Zhengkai Jiang, Muyang He, Bo~Zhao, Xin Tan, Zhenye Gan, et~al.
\newblock Efficient multimodal large language models: A survey.
\newblock \emph{arXiv preprint arXiv:2405.10739}, 2024.

\bibitem[Khanam \& Hussain(2024)Khanam and Hussain]{yolov11}
Rahima Khanam and Muhammad Hussain.
\newblock Yolov11: An overview of the key architectural enhancements.
\newblock \emph{arXiv preprint arXiv:2410.17725}, 2024.

\bibitem[Lauren{\c{c}}on et~al.(2024)Lauren{\c{c}}on, Marafioti, Sanh, and Tronchon]{idfecs}
Hugo Lauren{\c{c}}on, Andr{\'e}s Marafioti, Victor Sanh, and L{\'e}o Tronchon.
\newblock Building and better understanding vision-language models: insights and future directions.
\newblock \emph{arXiv preprint arXiv:2408.12637}, 2024.

\bibitem[Leng et~al.(2024)Leng, Zhang, Chen, Li, Lu, Miao, and Bing]{VCD}
Sicong Leng, Hang Zhang, Guanzheng Chen, Xin Li, Shijian Lu, Chunyan Miao, and Lidong Bing.
\newblock Mitigating object hallucinations in large vision-language models through visual contrastive decoding.
\newblock In \emph{Proceedings of the IEEE/CVF Conference on Computer Vision and Pattern Recognition}, pp.\  13872--13882, 2024.

\bibitem[Lerner et~al.(2022)Lerner, Ferret, Guinaudeau, Le~Borgne, Besan{\c{c}}on, Moreno, and Lov{\'o}n~Melgarejo]{Viquae}
Paul Lerner, Olivier Ferret, Camille Guinaudeau, Herv{\'e} Le~Borgne, Romaric Besan{\c{c}}on, Jos{\'e}~G Moreno, and Jes{\'u}s Lov{\'o}n~Melgarejo.
\newblock Viquae, a dataset for knowledge-based visual question answering about named entities.
\newblock In \emph{Proceedings of the 45th international ACM SIGIR conference on research and development in information retrieval}, pp.\  3108--3120, 2022.

\bibitem[Li et~al.(2024{\natexlab{a}})Li, Ge, Ge, Wang, Wang, Zhang, and Shan]{seed}
Bohao Li, Yuying Ge, Yixiao Ge, Guangzhi Wang, Rui Wang, Ruimao Zhang, and Ying Shan.
\newblock Seed-bench: Benchmarking multimodal large language models.
\newblock In \emph{Proceedings of the IEEE/CVF Conference on Computer Vision and Pattern Recognition}, pp.\  13299--13308, 2024{\natexlab{a}}.

\bibitem[Li et~al.(2023{\natexlab{a}})Li, Li, Savarese, and Hoi]{blip2}
Junnan Li, Dongxu Li, Silvio Savarese, and Steven Hoi.
\newblock Blip-2: Bootstrapping language-image pre-training with frozen image encoders and large language models.
\newblock In \emph{International conference on machine learning}, pp.\  19730--19742. PMLR, 2023{\natexlab{a}}.

\bibitem[Li et~al.(2023{\natexlab{b}})Li, Patel, Vi{\'e}gas, Pfister, and Wattenberg]{ITI}
Kenneth Li, Oam Patel, Fernanda Vi{\'e}gas, Hanspeter Pfister, and Martin Wattenberg.
\newblock Inference-time intervention: Eliciting truthful answers from a language model.
\newblock \emph{Advances in Neural Information Processing Systems}, 36:\penalty0 41451--41530, 2023{\natexlab{b}}.

\bibitem[Li et~al.(2024{\natexlab{b}})Li, Zhang, Geng, Geng, Long, Shen, Zhang, Liu, and Dong]{robotic}
Xiaoqi Li, Mingxu Zhang, Yiran Geng, Haoran Geng, Yuxing Long, Yan Shen, Renrui Zhang, Jiaming Liu, and Hao Dong.
\newblock Manipllm: Embodied multimodal large language model for object-centric robotic manipulation.
\newblock In \emph{Proceedings of the IEEE/CVF Conference on Computer Vision and Pattern Recognition}, pp.\  18061--18070, 2024{\natexlab{b}}.

\bibitem[Li et~al.(2023{\natexlab{c}})Li, Du, Zhou, Wang, Zhao, and Wen]{POPE}
Yifan Li, Yifan Du, Kun Zhou, Jinpeng Wang, Wayne~Xin Zhao, and Ji-Rong Wen.
\newblock Evaluating object hallucination in large vision-language models.
\newblock In \emph{Proceedings of the 2023 Conference on Empirical Methods in Natural Language Processing}, pp.\  292--305, 2023{\natexlab{c}}.

\bibitem[Lin et~al.(2014)Lin, Maire, Belongie, Hays, Perona, Ramanan, Doll{\'a}r, and Zitnick]{coco}
Tsung-Yi Lin, Michael Maire, Serge Belongie, James Hays, Pietro Perona, Deva Ramanan, Piotr Doll{\'a}r, and C~Lawrence Zitnick.
\newblock Microsoft coco: Common objects in context.
\newblock In \emph{Computer vision--ECCV 2014: 13th European conference, zurich, Switzerland, September 6-12, 2014, proceedings, part v 13}, pp.\  740--755. Springer, 2014.

\bibitem[Liu et~al.(2024{\natexlab{a}})Liu, Lin, Li, Wang, Yacoob, and Wang]{LRV}
Fuxiao Liu, Kevin Lin, Linjie Li, Jianfeng Wang, Yaser Yacoob, and Lijuan Wang.
\newblock Mitigating hallucination in large multi-modal models via robust instruction tuning.
\newblock In \emph{The Twelfth International Conference on Learning Representations}, 2024{\natexlab{a}}.

\bibitem[Liu et~al.(2024{\natexlab{b}})Liu, Xue, Chen, Chen, Zhao, Wang, Hou, Li, and Peng]{lvlmhallusurvey}
Hanchao Liu, Wenyuan Xue, Yifei Chen, Dapeng Chen, Xiutian Zhao, Ke~Wang, Liping Hou, Rongjun Li, and Wei Peng.
\newblock A survey on hallucination in large vision-language models.
\newblock \emph{arXiv preprint arXiv:2402.00253}, 2024{\natexlab{b}}.

\bibitem[Liu et~al.(2023)Liu, Li, Wu, and Lee]{llava}
Haotian Liu, Chunyuan Li, Qingyang Wu, and Yong~Jae Lee.
\newblock Visual instruction tuning.
\newblock \emph{Advances in neural information processing systems}, 36:\penalty0 34892--34916, 2023.

\bibitem[Liu et~al.(2024{\natexlab{c}})Liu, Li, Li, and Lee]{llava1.5}
Haotian Liu, Chunyuan Li, Yuheng Li, and Yong~Jae Lee.
\newblock Improved baselines with visual instruction tuning.
\newblock In \emph{Proceedings of the IEEE/CVF Conference on Computer Vision and Pattern Recognition}, pp.\  26296--26306, 2024{\natexlab{c}}.

\bibitem[Liu et~al.(2025{\natexlab{a}})Liu, Fu, Xie, Xie, Sun, Lian, Kang, and Li]{phd}
Jiazhen Liu, Yuhan Fu, Ruobing Xie, Runquan Xie, Xingwu Sun, Fengzong Lian, Zhanhui Kang, and Xirong Li.
\newblock Phd: A chatgpt-prompted visual hallucination evaluation dataset.
\newblock In \emph{Proceedings of the Computer Vision and Pattern Recognition Conference}, pp.\  19857--19866, 2025{\natexlab{a}}.

\bibitem[Liu et~al.(2025{\natexlab{b}})Liu, Ye, and Zou]{VTI}
Sheng Liu, Haotian Ye, and James Zou.
\newblock Reducing hallucinations in large vision-language models via latent space steering.
\newblock In \emph{The Thirteenth International Conference on Learning Representations}, 2025{\natexlab{b}}.

\bibitem[Lu et~al.(2022)Lu, Mishra, Xia, Qiu, Chang, Zhu, Tafjord, Clark, and Kalyan]{scienceqa}
Pan Lu, Swaroop Mishra, Tanglin Xia, Liang Qiu, Kai-Wei Chang, Song-Chun Zhu, Oyvind Tafjord, Peter Clark, and Ashwin Kalyan.
\newblock Learn to explain: Multimodal reasoning via thought chains for science question answering.
\newblock \emph{Advances in Neural Information Processing Systems}, 35:\penalty0 2507--2521, 2022.

\bibitem[Radford et~al.(2021)Radford, Kim, Hallacy, Ramesh, Goh, Agarwal, Sastry, Askell, Mishkin, Clark, et~al.]{clip}
Alec Radford, Jong~Wook Kim, Chris Hallacy, Aditya Ramesh, Gabriel Goh, Sandhini Agarwal, Girish Sastry, Amanda Askell, Pamela Mishkin, Jack Clark, et~al.
\newblock Learning transferable visual models from natural language supervision.
\newblock In \emph{International conference on machine learning}, pp.\  8748--8763. PMLR, 2021.

\bibitem[Rohrbach et~al.(2018)Rohrbach, Hendricks, Burns, Darrell, and Saenko]{Chair}
Anna Rohrbach, Lisa~Anne Hendricks, Kaylee Burns, Trevor Darrell, and Kate Saenko.
\newblock Object hallucination in image captioning.
\newblock In \emph{Proceedings of the 2018 Conference on Empirical Methods in Natural Language Processing}, pp.\  4035--4045, 2018.

\bibitem[Shahgir et~al.(2024)Shahgir, Sayeed, Bhattacharjee, Ahmad, Dong, and Shahriyar]{illusionvqa}
Haz~Sameen Shahgir, Khondker~Salman Sayeed, Abhik Bhattacharjee, Wasi~Uddin Ahmad, Yue Dong, and Rifat Shahriyar.
\newblock Illusionvqa: A challenging optical illusion dataset for vision language models.
\newblock \emph{arXiv preprint arXiv:2403.15952}, 2024.

\bibitem[Su et~al.(2023)Su, Lan, Li, Xu, Wang, and Cai]{pandagpt}
Yixuan Su, Tian Lan, Huayang Li, Jialu Xu, Yan Wang, and Deng Cai.
\newblock Pandagpt: One model to instruction-follow them all.
\newblock In \emph{Proceedings of the 1st Workshop on Taming Large Language Models: Controllability in the era of Interactive Assistants!}, pp.\  11--23, 2023.

\bibitem[Tong et~al.(2024)Tong, Liu, Zhai, Ma, LeCun, and Xie]{MMVP}
Shengbang Tong, Zhuang Liu, Yuexiang Zhai, Yi~Ma, Yann LeCun, and Saining Xie.
\newblock Eyes wide shut? exploring the visual shortcomings of multimodal llms.
\newblock In \emph{Proceedings of the IEEE/CVF Conference on Computer Vision and Pattern Recognition}, pp.\  9568--9578, 2024.

\bibitem[Wang et~al.(2024{\natexlab{a}})Wang, Chen, Zhang, Tian, Xu, Deng, and Chen]{deco}
Chenxi Wang, Xiang Chen, Ningyu Zhang, Bozhong Tian, Haoming Xu, Shumin Deng, and Huajun Chen.
\newblock Mllm can see? dynamic correction decoding for hallucination mitigation.
\newblock \emph{arXiv preprint arXiv:2410.11779}, 2024{\natexlab{a}}.

\bibitem[Wang et~al.(2023)Wang, Wang, Xu, Zhang, Gu, Jia, Wang, Xu, Yan, Zhang, et~al.]{amber}
Junyang Wang, Yuhang Wang, Guohai Xu, Jing Zhang, Yukai Gu, Haitao Jia, Jiaqi Wang, Haiyang Xu, Ming Yan, Ji~Zhang, et~al.
\newblock Amber: An llm-free multi-dimensional benchmark for mllms hallucination evaluation.
\newblock \emph{arXiv preprint arXiv:2311.07397}, 2023.

\bibitem[Wang et~al.(2025)Wang, Gu, Gao, and Zhou]{damo}
Kaishen Wang, Hengrui Gu, Meijun Gao, and Kaixiong Zhou.
\newblock Damo: Decoding by accumulating activations momentum for mitigating hallucinations in vision-language models.
\newblock In \emph{The Thirteenth International Conference on Learning Representations}, 2025.

\bibitem[Wang et~al.(2024{\natexlab{b}})Wang, Yang, and Peng]{SADI}
Weixuan Wang, Jingyuan Yang, and Wei Peng.
\newblock Semantics-adaptive activation intervention for llms via dynamic steering vectors.
\newblock \emph{arXiv preprint arXiv:2410.12299}, 2024{\natexlab{b}}.

\bibitem[Wang et~al.(2024{\natexlab{c}})Wang, Pan, Ding, and Biemann]{ICD}
Xintong Wang, Jingheng Pan, Liang Ding, and Chris Biemann.
\newblock Mitigating hallucinations in large vision-language models with instruction contrastive decoding.
\newblock In \emph{Findings of the Association for Computational Linguistics ACL 2024}, pp.\  15840--15853, 2024{\natexlab{c}}.

\bibitem[Yin et~al.(2025)Yin, Si, and Wang]{VAF}
Hao Yin, Guangzong Si, and Zilei Wang.
\newblock Clearsight: Visual signal enhancement for object hallucination mitigation in multimodal large language models.
\newblock In \emph{Proceedings of the Computer Vision and Pattern Recognition Conference}, pp.\  14625--14634, 2025.

\bibitem[Yin et~al.(2023)Yin, Fu, Zhao, Li, Sun, Xu, and Chen]{MMsurvey1}
Shukang Yin, Chaoyou Fu, Sirui Zhao, Ke~Li, Xing Sun, Tong Xu, and Enhong Chen.
\newblock A survey on multimodal large language models.
\newblock \emph{arXiv preprint arXiv:2306.13549}, 2023.

\bibitem[Yin et~al.(2024)Yin, Fu, Zhao, Xu, Wang, Sui, Shen, Li, Sun, and Chen]{woodpecker}
Shukang Yin, Chaoyou Fu, Sirui Zhao, Tong Xu, Hao Wang, Dianbo Sui, Yunhang Shen, Ke~Li, Xing Sun, and Enhong Chen.
\newblock Woodpecker: Hallucination correction for multimodal large language models.
\newblock \emph{Science China Information Sciences}, 67\penalty0 (12):\penalty0 220105, 2024.

\bibitem[Yu et~al.(2024{\natexlab{a}})Yu, Yao, Zhang, He, Han, Cui, Hu, Liu, Zheng, Sun, et~al.]{rlhf}
Tianyu Yu, Yuan Yao, Haoye Zhang, Taiwen He, Yifeng Han, Ganqu Cui, Jinyi Hu, Zhiyuan Liu, Hai-Tao Zheng, Maosong Sun, et~al.
\newblock Rlhf-v: Towards trustworthy mllms via behavior alignment from fine-grained correctional human feedback.
\newblock In \emph{Proceedings of the IEEE/CVF Conference on Computer Vision and Pattern Recognition}, pp.\  13807--13816, 2024{\natexlab{a}}.

\bibitem[Yu et~al.(2024{\natexlab{b}})Yu, Zhang, Li, Xu, Yao, Chen, Lu, Cui, Dang, He, Feng, Song, Zheng, Liu, Chua, and Sun]{RLAIF}
Tianyu Yu, Haoye Zhang, Qiming Li, Qixin Xu, Yuan Yao, Da~Chen, Xiaoman Lu, Ganqu Cui, Yunkai Dang, Taiwen He, Xiaocheng Feng, Jun Song, Bo~Zheng, Zhiyuan Liu, Tat-Seng Chua, and Maosong Sun.
\newblock Rlaif-v: Open-source ai feedback leads to super gpt-4v trustworthiness, 2024{\natexlab{b}}.
\newblock URL \url{https://arxiv.org/abs/2405.17220}.

\bibitem[Yu et~al.(2024{\natexlab{c}})Yu, Yang, Li, Wang, Lin, Liu, Wang, and Wang]{mmvet}
Weihao Yu, Zhengyuan Yang, Linjie Li, Jianfeng Wang, Kevin Lin, Zicheng Liu, Xinchao Wang, and Lijuan Wang.
\newblock Mm-vet: evaluating large multimodal models for integrated capabilities.
\newblock In \emph{Proceedings of the 41st International Conference on Machine Learning}, pp.\  57730--57754, 2024{\natexlab{c}}.

\bibitem[Zhang et~al.(2025)Zhang, Wan, Kan, Ma, Stepputtis, Ramanan, Salakhutdinov, Morency, Sycara, and Xie]{DeFG}
Ce~Zhang, Zifu Wan, Zhehan Kan, Martin~Q Ma, Simon Stepputtis, Deva Ramanan, Russ Salakhutdinov, Louis-Philippe Morency, Katia~P Sycara, and Yaqi Xie.
\newblock Self-correcting decoding with generative feedback for mitigating hallucinations in large vision-language models.
\newblock In \emph{The Thirteenth International Conference on Learning Representations}, 2025.

\bibitem[Zhao et~al.(2025)Zhao, Zhang, Sun, and Feng]{CICD}
Jianfei Zhao, Feng Zhang, Xin Sun, and Chong Feng.
\newblock Mitigate language priors in large vision-language models by cross-images contrastive decoding.
\newblock \emph{arXiv e-prints}, pp.\  arXiv--2505, 2025.

\bibitem[Zhou et~al.(2024)Zhou, Cui, Yoon, Zhang, Deng, Finn, Bansal, and Yao]{LURE}
Yiyang Zhou, Chenhang Cui, Jaehong Yoon, Linjun Zhang, Zhun Deng, Chelsea Finn, Mohit Bansal, and Huaxiu Yao.
\newblock Analyzing and mitigating object hallucination in large vision-language models.
\newblock In \emph{The Twelfth International Conference on Learning Representations}, 2024.

\bibitem[Zhu et~al.(2025{\natexlab{a}})Zhu, Wang, Chen, Liu, Ye, Gu, Tian, Duan, Su, Shao, et~al.]{zhu2025internvl3}
Jinguo Zhu, Weiyun Wang, Zhe Chen, Zhaoyang Liu, Shenglong Ye, Lixin Gu, Hao Tian, Yuchen Duan, Weijie Su, Jie Shao, et~al.
\newblock Internvl3: Exploring advanced training and test-time recipes for open-source multimodal models.
\newblock \emph{arXiv preprint arXiv:2504.10479}, 2025{\natexlab{a}}.

\bibitem[Zhu et~al.(2025{\natexlab{b}})Zhu, Ji, Chen, Xu, Ye, and Liu]{ibd}
Lanyun Zhu, Deyi Ji, Tianrun Chen, Peng Xu, Jieping Ye, and Jun Liu.
\newblock Ibd: Alleviating hallucinations in large vision-language models via image-biased decoding.
\newblock In \emph{Proceedings of the Computer Vision and Pattern Recognition Conference}, pp.\  1624--1633, 2025{\natexlab{b}}.

\bibitem[Zou et~al.(2023)Zou, Phan, Chen, Campbell, Guo, Ren, Pan, Yin, Mazeika, Dombrowski, et~al.]{representation}
Andy Zou, Long Phan, Sarah Chen, James Campbell, Phillip Guo, Richard Ren, Alexander Pan, Xuwang Yin, Mantas Mazeika, Ann-Kathrin Dombrowski, et~al.
\newblock Representation engineering: A top-down approach to ai transparency.
\newblock \emph{arXiv preprint arXiv:2310.01405}, 2023.

\end{thebibliography}
% \bibliographystyle{iclr2026_conference}

\clearpage
\appendix
\section{Appendix}
\subsection{AI Writing Assistance Statement}
Large language models were utilized solely for minor linguistic improvements, including enhanced phrasing and clarity. These tools played no role in content generation, experimental design, data analysis, or interpretation. The authors are entirely responsible for all ideas, results, and conclusions presented in this paper.

\subsection{More Details on CHAIR}

In this paper, we report $\text{CHAIR}_S$ and $\text{CHAIR}_I$ as evaluation metrics. The calculation of $\text{CHAIR}_S$ and $\text{CHAIR}_I$ is shown in Equation \ref{chair}, where we set the maximum number of new tokens to 512 in our experiments.

\begin{equation}
\begin{aligned}
\text{CHAIR}_S &= \frac{|\{\text{sentences with hallucinated objects}\}|}{|\{\text{all sentences}\}|} \\
\text{CHAIR}_I &= \frac{|\{\text{hallucinated objects}\}|}{|\{\text{all objects mentioned}\}|}
\end{aligned}
\label{chair}
\end{equation}

\subsection{Results on POPE}
The complete experimental results on POPE are presented in Table \ref{tab:POPE_llava}. Our method achieves significant improvements across all three experimental settings: random, popular, and adversarial.

\begin{table*}[thb!]
  \centering
  \begin{tabular}{ccccccccc}
    \toprule
    Dataset& Setting & Method & Accuracy $\uparrow$ & Precision & Recall & F1 Score $\uparrow$ \\
    \midrule
\multirow{9}{*}{MSCOCO} & \multirow{3}{*}{Random} & Regular & 83.29	& 92.13	& 72.80	&81.33 \\
                              &                    & VCD & 87.73	&91.42	&83.28	&87.16 \\
                             &                    &  Ours & \textbf{90.03}	&90.51	&90.03	&\textbf{90.02} \\
\cmidrule{2-7}
                              & \multirow{3}{*}{Popular} & Regular & 81.88	&88.93 	&72.80	&80.06 \\
                              &                    & VCD & 85.38	&86.92	&83.28	&85.06\\\
                             &                    &  Ours & \textbf{87.33}	&89.16	&85.00	&\textbf{87.03} \\ 
\cmidrule{2-7}
                              & \multirow{3}{*}{Adversarial} & Regular & 78.96	&83.06 	&72.75	&77.57 \\
                              &                    & VCD & 80.88	&79.45	&83.29	&81.33\\
                             &                    & Ours & \textbf{83.07}	&82.04	&84.67	&\textbf{83.33} \\ 
\midrule
\multirow{9}{*}{GQA}  & \multirow{3}{*}{Random} & Regular & 83.73 	&87.16	&79.12	&82.95\\
                              &                    & VCD & 86.65	&84.85	&89.24	&86.99 \\
                             &                    &  Ours & \textbf{89.57}	&88.92	&90.40	&\textbf{89.60} \\ 
\cmidrule{2-7}                              
                              & \multirow{3}{*}{Popular} & Regular & 78.17	&77.64	&79.12	&78.37 \\
                              &                    & VCD & 80.73	&76.26	&89.24	&82.24 \\
                             &                    &  Ours & \textbf{84.53}	&83.51	&86.07	&\textbf{84.77} \\ 
\cmidrule{2-7}                              
                              & \multirow{3}{*}{Adversarial} & Regular & 75.08	&73.19	&79.16	&76.06 \\
                              &                    & VCD & 76.09	&70.83	&88.75	&78.78 \\
                             &                    &  Ours & \textbf{81.70}	&79.15	&86.07	&\textbf{82.47} \\ 
    \bottomrule
  \end{tabular}
  \caption{Results on LLaVAv1.5. The best results are shown in bold.}
  \label{tab:POPE_llava}
\end{table*}

\subsection{Results on AMBER}
We conduct an evaluation of LLaVA v1.5 on the AMBER \citep{amber}. AMBER contains both discriminative tasks and generative tasks. The experimental results are shown in Table \ref{tab:amber}. Our method outperforms existing methods on both discriminative and generative tasks, achieving significant effects in hallucination mitigation.

\begin{table*}[htbp]
\centering
\begin{tabular}{lcccccc}
\toprule
& \multicolumn{2}{c}{Discriminative} & \multicolumn{2}{c}{Generative} & \\
\cmidrule(lr){2-3} \cmidrule(lr){4-5}
Method & Acc$\uparrow$ & F1$\uparrow$ & CHAIR$\downarrow$ & Hal$\downarrow$ & AMBER SCORE$\uparrow$ \\
\midrule
Regular & 67.4 & 71.2 & 11.6 & 47.7 & 79.80 \\
VCD \citep{VCD} & 68.1 & 71.1 & 9.8 & 43.8 & 80.65 \\
ICD \citep{ICD} & 70.3 & 73.4 & 8.8 & 38.7 & 82.3 \\
IBD \citep{ibd} & 69.2 & 72.2 & 9.8 & 42.2 & 81.2 \\
DeFG \citep{DeFG} & 70.2 & 73.0 & 9.1 & 39.9 & 81.95 \\
CICD \citep{CICD} & 71.1 & 73.1 & 6.6 & 34.8 & 83.25 \\
Ours & \textbf{81.9} & \textbf{87.2} & \textbf{4.9} & \textbf{20.9} & \textbf{90.01} \\
\bottomrule
\end{tabular}
\caption{Results on AMBER. The best results are shown in bold.}
\label{tab:amber}
\end{table*}

\subsection{Scalability Analysis Across Model Sizes}
To verify that our method has hallucination mitigation effects for models of different sizes, we select the discriminative task dataset MME and the generative task dataset CHAIR on LLaVAv1.5 7B and 13B models. The experimental results show in Table \ref{tab:size} that our model achieves significant effects for models of different sizes in both discriminative and generative tasks.

\begin{table}[thb!]
  \centering
  \scalebox{0.8}{
    \begin{tabular}{cccccccccc}
      \toprule
      & & \multicolumn{5}{c}{MME} & \multicolumn{2}{c}{CHAIR} \\
      \cmidrule(lr){3-7} \cmidrule(lr){8-9}
      Model& Method & Existence$\uparrow$ & Count $\uparrow$& Position$\uparrow$& Color$\uparrow$ & Total Scores$\uparrow$  &$CHAIR_S$$\downarrow$ &  $CHAIR_I$$\downarrow$\\
      \midrule
      \multirow{2}{*}{LLaVAv1.5 7B}     & Regular & 175.67 & 124.67 & 114.00 & 151.00 & 565.33 &51.0 &15.2\\
                               & Ours & \textbf{195.00}   & \textbf{158.33} & \textbf{133.33} & \textbf{173.33} & \textbf{659.99}&\textbf{30.8}&\textbf{11.4}\\
      \midrule
      \multirow{2}{*}{LLaVA1.5 13B}  & Regular & \textbf{185.00} & 131.67 & 95.00 & 175.00 & 586.67 &45.0 &11.8\\
                           & Ours & \textbf{185.00} & \textbf{158.33} & \textbf{103.33} & \textbf{180.00} & \textbf{626.66} &\textbf{38.0}&\textbf{10.8}\\
      \bottomrule
    \end{tabular}
  }
  \caption{Scalability Analysis Across Model Sizes on MME and CHAIR. The best results are shown in bold.}
  \label{tab:size}
\end{table}

\subsection{Analysis of Inference Speed}
\begin{figure}[]
  \centering
  \includegraphics[width=0.5\textwidth]{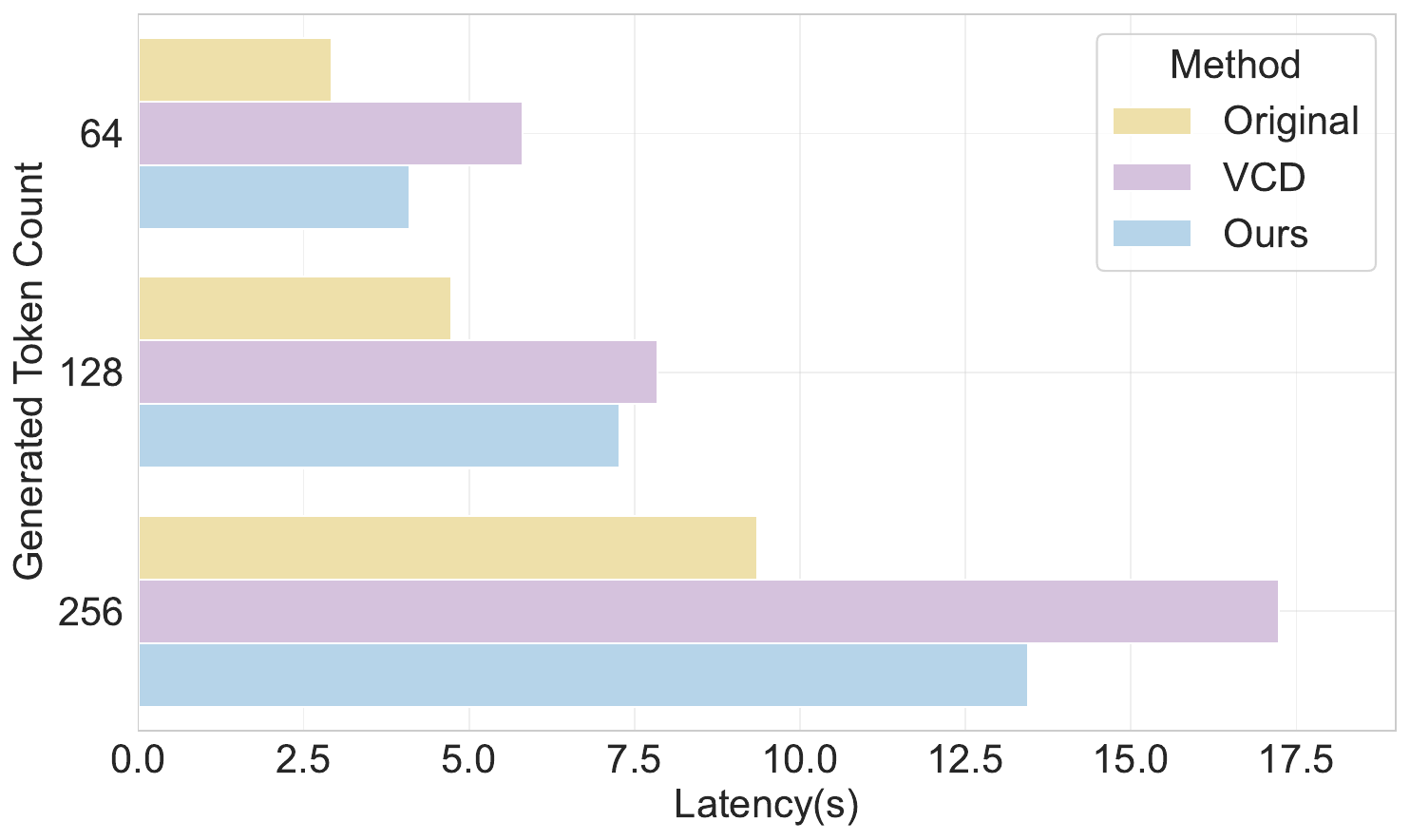}\\
  \caption{Effect of dynamic intervention.}
  \label{speed}
\end{figure}

In this section, we investigate DMAS's inference speed. We use LLaVAv1.5 7B and set the generation content lengths to \{64, 128, 256\} respectively, then compare the inference speed of our method with the original model and VCD. The experimental results show in Figure \ref{speed} that our method has faster inference speed compared to the decoding method VCD. VCD's inference latency is almost twice that of the original model, but our model achieves better hallucination mitigation effects while adding only a small amount of inference time.

\subsection{Case Study}
\begin{figure}[thb!]
  \centering
  \includegraphics[width=1\textwidth]{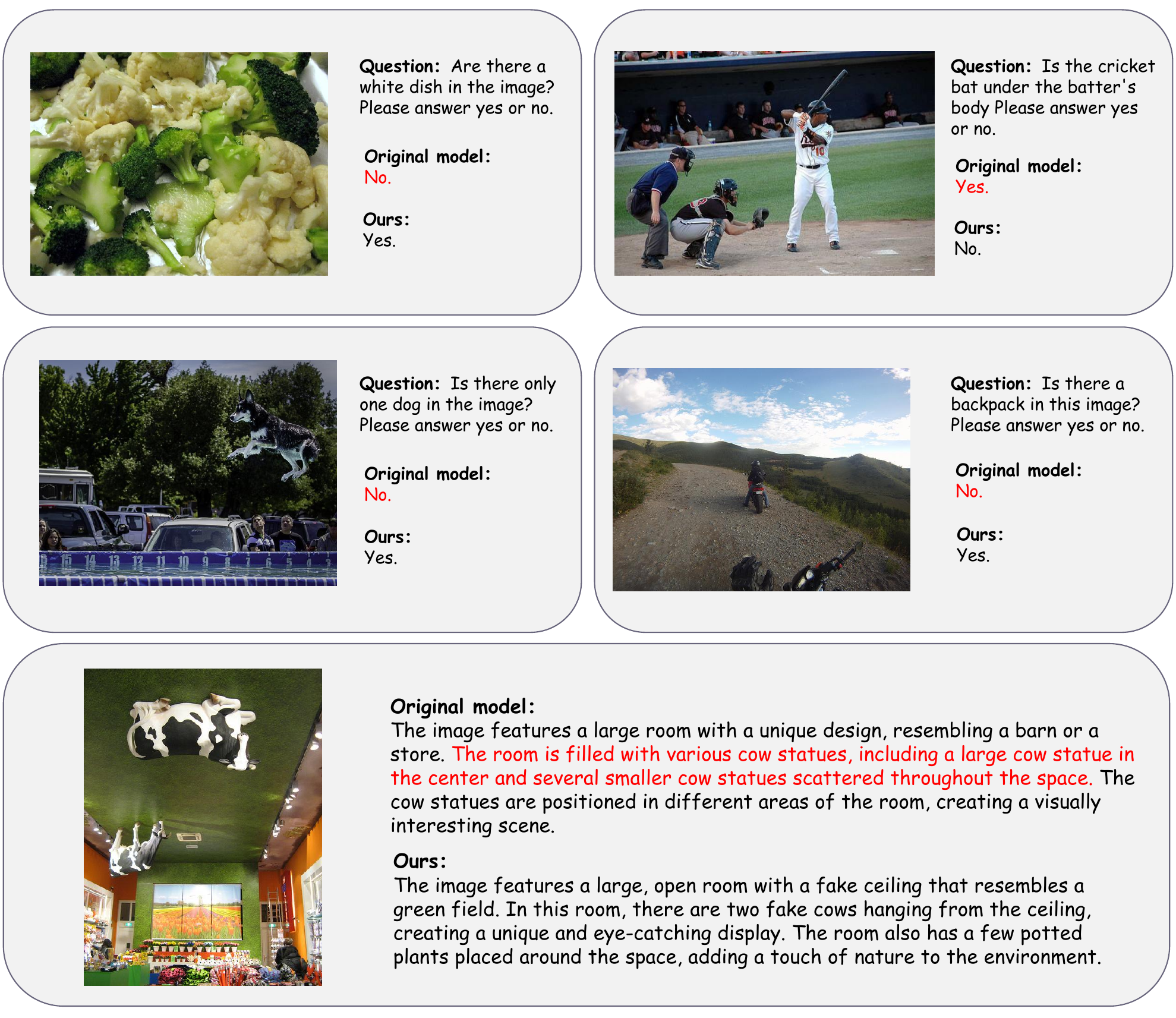}\\
  \caption{Case study on LLaVAv1.5.}
  \label{case}
\end{figure}

To intuitively demonstrate the hallucination mitigation effectiveness of our method, we conduct case studies on LLaVAv1.5. We utilize cases from MME and CHAIR datasets, with results shown in Figure \ref{case}. Our method effectively mitigates multimodal hallucination issues across both VQA tasks and image captioning tasks. For VQA tasks, we present various question types, demonstrating our method's effectiveness in reducing hallucinations at different levels including object, attribute, relation, and count. For image captioning tasks, our method not only generates fewer hallucinations but also maintains the quality of the output content.

\subsection{Experiments on more benchmarks}

To further substantiate the effectiveness of our approach, we conduct extensive experiments across multiple challenging benchmarks, including MMVP \citep{MMVP}, IllusionVQA \citep{illusionvqa}, HallusionBench (multimodal subset) \citep{hallusionbench}, and MM-Vet \citep{mmvet}, evaluating on both LLaVAv1.5 and QwenVL. The results summarized in the Table \ref{tab:more bench} confirm that our method delivers strong and robust performance across these diverse and complex evaluations.

\begin{table}[thb!]
  \centering
  \scalebox{1}{
    \begin{tabular}{cccccc}
      \toprule
       Model&Method & MMVP & MM-Vet & IllusionVQA & HallusionBench \\
      \midrule
      \multirow{3}{*}{LLaVAv1.5}     & Vanilla & 30.3 & 38.9 & 32.6 & 50.2 \\
                            &ICT &31.0 &33.4 &33.1	&52.7 \\
                              & Ours & \textbf{54.3}   & \textbf{39.3} & \textbf{35.2} & \textbf{53.7} \\
      \midrule
      \multirow{2}{*}{QwenVL}  & Vanilla & 41.0 & 32.9 & 31.7 & 42.4 \\
                          & Ours &\textbf{ 43.3} & \textbf{34.4} & \textbf{33.6} & \textbf{42.9} \\
      \bottomrule
    \end{tabular}
  }
  \caption{Experiments on more benchmarks.}
  \label{tab:more bench}
\end{table}

\subsection{Comprehensive analysis of the constructed dataset}

In this subsection, we will explore how the dataset scale and the dataset composition for constructing steering vectors respectively affect the effectiveness of DMAS.
\subsubsection{Dataset Scale}
To investigate the impact of dataset size on steering vector construction, we systematically reduce the dataset to 75\%, 50\%, and 25\% of its original size and evaluated performance on MME \citep{MME}, AMBER \citep{amber}, and MMVP \citep{MMVP}. The results are presented in Table \ref{tab:data scale}. The results show that dataset size has minimal impact on performance, with only minor fluctuations across benchmarks. This demonstrates that our method is robust to variations in dataset scale and does not require large amounts of labeled data.
\begin{table}[thb!]
  \centering
  \scalebox{1}{
    \begin{tabular}{ccccccccc}
      \toprule
      & \multicolumn{4}{c}{MME} & MMVP &\multicolumn{2}{c}{AMBER} \\
      \cmidrule(lr){2-5} \cmidrule(lr){7-8}
      Scale & Existence$\uparrow$ & Count $\uparrow$& Position$\uparrow$& Color$\uparrow$ & Accuracy$\uparrow$  & Relation$\uparrow$ &  F1 $\uparrow$\\
      \midrule
      100\%      &195.00	&158.33	&133.33	&173.33	&54.3	&69.0	&87.2\\

      75\%  &195.00	&158.33	&133.33	&168.33	&52.0	&69.0	&87.2\\

      50\%   &195.00	&158.33	&133.33	&168.33	&52.0	&69.0	&87.2\\

      25\%   &195.00	&158.33	&133.33	&168.33	&53.0	&69.0	&87.1\\

      \bottomrule
    \end{tabular}
  }
  \caption{Analysis of dataset scale for steering vectors construction.}
  \label{tab:data scale}
\end{table}
\subsubsection{Dataset Composition}

To explore how different datasets affect steering vector construction, we build databases using two alternative datasets: POPE \citep{POPE} and PHD \citep{phd}. POPE contains only object existence questions (homogeneous), while PHD includes diverse question types similar to our original datasets. Both databases are constructed using the same scale as our main experiments, and we evaluate performance on MME \citep{MME}, AMBER \citep{amber}, and MMVP \citep{MMVP}. The results are presented in Table \ref{tab:data composition}.
Results show that the POPE-based database underperforms on several MME subtasks and AMBER's relation subtask, which is expected given POPE's limited diversity cannot fully exploit our dynamic approach. In contrast, the PHD-based database achieves comparable results to our original method across all benchmarks. These findings demonstrate two important insights: (1) when using datasets with good question diversity, our method achieves consistent performance regardless of the specific dataset chosen, and (2) this validates the necessity of our dynamic retrieval mechanism for selecting appropriate steering vectors based on semantic context.

\begin{table}[thb!]
  \centering
  \scalebox{1}{
    \begin{tabular}{ccccccccc}
      \toprule
      & \multicolumn{4}{c}{MME} & MMVP &\multicolumn{2}{c}{AMBER} \\
      \cmidrule(lr){2-5} \cmidrule(lr){7-8}
      Scale & Existence$\uparrow$ & Count $\uparrow$& Position$\uparrow$& Color$\uparrow$ & Accuracy$\uparrow$  & Relation$\uparrow$ &  F1 $\uparrow$\\
      \midrule
      Ours      &195.00	&158.33	&133.33	&173.33	&54.3	&69.0	&87.2\\

      POPE  &195.00	&158.33	&123.33	&153.33	&52.3	&65.1	&86.0\\

      PHD   &195.00	&158.33	&133.33	&173.33	&53.7	&68.3	&86.9\\

      \bottomrule
    \end{tabular}
  }
  \caption{Analysis of dataset composition for steering vectors construction.}
  \label{tab:data composition}
\end{table}

\subsection{Experiments on different decoding strategies}

To explore the effectiveness of our method under different decoding strategies, including beam search (beam=2), nucleus sampling (temperature=0.5, topp=0.7) and greedy search, the results on MME and MMVP benchmarks are presented in the Table \ref{tab:decoding}. Our method consistently demonstrates effective hallucination mitigation across all decoding settings, confirming the robustness of our approach.

\begin{table}[thb!]
  \centering
  \scalebox{0.9}{
    \begin{tabular}{cccccccccc}
      \toprule
      & & \multicolumn{5}{c}{MME} & MMVP \\
      \cmidrule(lr){3-7}
      Decoding Strategies& Method & Existence$\uparrow$ & Count $\uparrow$& Position$\uparrow$& Color$\uparrow$ & Total Scores$\uparrow$  &Accuracy$\uparrow$\\
      \midrule
      \multirow{2}{*}{Beam}     & Vanilla & 190.00	&153.33	&113.33	&148.33	&604.99	&29.7\\
                               & Ours  &\textbf{191.67} 	&\textbf{156.66} 	&\textbf{135.00} 	&\textbf{171.66} 	&\textbf{654.99}	&\textbf{55.1}\\
      \midrule
      \multirow{2}{*}{Nucleus}  & Vanilla & 190.00	&147.78	&123.89	&150.55	&612.22	&32.3\\
                           & Ours & \textbf{190.00}	&\textbf{150.00}	&\textbf{130.00}	&\textbf{159.44}	&\textbf{629.44}	&\textbf{53.1}\\
              \midrule
        \multirow{2}{*}{Greedy}  & Vanilla & 190.00	&153.33	&113.33	&148.33	&604.99	&30.3\\
                           & Ours & \textbf{195.00}	&\textbf{158.33}	&\textbf{133.33}	&\textbf{173.33}	&\textbf{659.99}	&\textbf{54.3}\\
      \bottomrule
    \end{tabular}
  }
  \caption{Experiments on different decoding strategies.}
  \label{tab:decoding}
\end{table}

\subsection{Experiments on more models}

In this subsection, we test the effectiveness of our method on several advanced LVLMs. We select Qwen2.5-VL \citep{qwen25vl}, InternVL3 \citep{zhu2025internvl3}, Mantis \citep{mantis}, and Idefics3 \citep{idfecs}, and conduct experiments on MME and CHAIR using greedy search. The experimental results are shown in Table \ref{tab:more models}. Our method remains effective for these advanced LVLMs with different architectures, demonstrating hallucination mitigation effects on both discriminative tasks and open-ended generation tasks, which confirms the scalability of our approach.

\begin{table}[thb!]
  \centering
  \scalebox{1}{
    \begin{tabular}{ccccc}
      \toprule
      & & MME & \multicolumn{2}{c}{CHAIR} \\
      \cmidrule(lr){4-5}

       Model&Method & Total Scores$\uparrow$  &$CHAIR_S$$\downarrow$ &  $CHAIR_I$$\downarrow$ \\
      \midrule
      \multirow{2}{*}{Qwen2.5-VL}     & Vanilla & 698.33	&47.4	&10 \\
                              & Ours & \textbf{718.33}	&\textbf{43.2}	&\textbf{9.5} \\
      \midrule
      \multirow{2}{*}{InternVL3}  & Vanilla & 698.33	&28.4	&7.6 \\
                          & Ours &\textbf{713.33}	&\textbf{26.4}	&\textbf{6.9} \\
    \midrule
    \multirow{2}{*}{Mantis}  & Vanilla & 653.33	&44.6	&12.7 \\
                          & Ours &\textbf{668.33}	&\textbf{34.8}	&\textbf{9.7} \\
        \midrule
        \multirow{2}{*}{Idefics3}  & Vanilla & 665.00	&49.9	&8.8 \\
                          & Ours &\textbf{678.33}	&\textbf{45.2}	&\textbf{8.2} \\
      \bottomrule
    \end{tabular}
  }
  \caption{Experiments on more models.}
  \label{tab:more models}
\end{table}

\subsection{Ablation studies on hallucination objects selection strategies}
For visual perception attention heads, we compare two hallucination objects selection strategies: (1) random selection across categories (e.g., people → panda), and (2) hard negatives within the same category (e.g., runner → swimmer). 
As shown in the Table \ref{tab:objects}, experiments on MME and CHAIR benchmarks demonstrate that both strategies achieve comparable performance with no significant difference. This indicates that our visual steering vector approach is robust to the choice of negative selection strategy.

\begin{table}[thb!]
  \centering
  \scalebox{0.8}{
    \begin{tabular}{ccccccccc}
      \toprule
      & \multicolumn{5}{c}{MME} &\multicolumn{2}{c}{CHAIR} \\
      \cmidrule(lr){2-6} \cmidrule(lr){7-8}
      Strategy & Existence$\uparrow$ & Count $\uparrow$& Position$\uparrow$& Color$\uparrow$ & Total Scores$\uparrow$  &$CHAIR_S$$\downarrow$ &  $CHAIR_I$$\downarrow$\\
      \midrule
      Random      &195.00	&158.33	&133.33	&173.33	&659.99	&31.2	&10.9\\

      Hard Negative  &195.00	&158.33	&133.33	&173.33	&659.99	&30.8	&11.4\\

      \bottomrule
    \end{tabular}
  }
  \caption{Ablation studies on hallucination objects selection strategies.}
  \label{tab:objects}
\end{table}

\end{document}